\documentclass{article} 
\usepackage{iclr2026_conference,times}

\usepackage{amsmath,amsfonts,bm}

\def\eqref#1{equation~\ref{#1}}

\def\1{\bm{1}}

\DeclareMathAlphabet{\mathsfit}{\encodingdefault}{\sfdefault}{m}{sl}
\SetMathAlphabet{\mathsfit}{bold}{\encodingdefault}{\sfdefault}{bx}{n}

\usepackage{hyperref}
\usepackage{url}
\usepackage{bbding}
\usepackage{bm}
\usepackage{float}
\usepackage{booktabs}
\usepackage{hyperref}
\usepackage{graphicx}
\usepackage{subfig}

\usepackage{tabu}                     
\usepackage{multirow}                 
\usepackage{multicol}                 
\usepackage{multirow}                
\usepackage{float}                    
\usepackage{makecell}                 
\usepackage{booktabs}   
\usepackage{colortbl}
\usepackage{wrapfig}
\usepackage{cleveref}

\title{ Rotation Control Unlearning: Quantifying and Controlling Continuous Unlearning for LLM with The Cognitive Rotation Space}

\author{Xiang Zhang$^{1}$, Kun Wei$^{1}$,Xu Yang$^{1}$, Jiahua Li$^{1}$, Su Yan$^{1}$, Cheng Deng$^{1}$\thanks{Corresponding author} \\
        $^{1}$ School of Electronic Engineering, Xidian University, Xi’an, Shaanxi, China, \\ 
         \texttt{\{zhangxiangxd,weikunsk,xuyang.xd,ljhxdu,chdeng.xd\}@gmail.com} \\
        \texttt{\{ys\}@stu.xidian.edu.cn}}

\iclrfinalcopy 
\begin{document}

\maketitle

\begin{abstract}

As Large Language Models (LLMs) become increasingly prevalent, their security vulnerabilities have already drawn attention.
Machine unlearning is introduced to seek to mitigate these risks by removing the influence of undesirable data. 
However, existing methods not only rely on the retained dataset to preserve model utility, but also suffer from cumulative catastrophic utility loss under continuous unlearning requests.
To solve this dilemma, we propose a novel method, called Rotation Control Unlearning (RCU), which leverages the rotational salience weight of RCU to quantify and control the unlearning degree in the continuous unlearning process.
The skew symmetric loss is designed to construct the existence of the cognitive rotation space, where the changes of rotational angle can simulate the continuous unlearning process.
Furthermore, we design an orthogonal rotation axes regularization to enforce mutually perpendicular rotation directions for continuous unlearning requests, effectively minimizing interference and addressing cumulative catastrophic utility loss.
Experiments on multiple datasets confirm that our method without retained dataset achieves SOTA performance.

\end{abstract}

\section{Introduction}
In recent years, the development of Large Language Models (LLMs) has received widespread attention. 
With the extensive application of GPT (\cite{gpt}) and other LLMs (\cite{deepseek,llama7B-2}) in academic research and industry (\cite{wang-survey}), concerns about LLMs have also increased. 
Among these, security issues regarding information protection have become particularly prominent.
These concerns have motivated researchers to use the machine unlearning method to remove potentially private (\cite{pan}), illegal or toxic data that may exist in LLMs.
Currently, machine unlearning in LLMs (\cite{bourtoule}) is mainly divided into two paradigms: the method based on parameters (\cite{canshuyouhua1,canshuyouhua2,canshuyouhua3-soul}) and the method based on in-context unlearning (\cite{shangxiawenxiezai1,shangxiawenxiezai2}). 
The methods based on parameters achieve effective unlearning by maximizing the task loss on the unlearning data (\cite{rethinking,falcon}).
The methods based on in-context unlearning modify the input prompts of LLM to make them refuse to output the content that needs to be unlearning (\cite{safety,unierase}).
Other methods achieve the unlearning goal by interfering with the LLM's representation of the unlearned data (\cite{5deep,backdoor}).

However, unlearning methods in LLM are often not a one-time operation but a continuous process in real world. 
Most of them exist the cumulative catastrophic utility loss (\cite{o3}) when dealing with continuous unlearning.
The cumulative catastrophic utility loss causes a significant decline in both the LLM's unlearning capability and utility retention capacity during the continuous unlearning process as the number of requests increases.
At the parameter level, this manifests as new unlearning requests inducing parameter shift in previously learned ones.
Furthermore, they still require a retained dataset to maintain the model's utility. 
This retained dataset consists of a part of the original training dataset (\cite{bourtoule}). 
Since LLM require a large amount of data for training (\cite{wang-survey}), using the retained dataset in continuous unlearning is not feasible (\cite{liu2025rethinking}).

The work of $o^3$ (\cite{o3}) proposes to mitigate cumulative catastrophic utility loss by imposing orthogonal constraints on LoRA parameters and introduces weights for the LoRA (\cite{lora}) modules to represent the degree of unlearning. 
However, this approach suffers from several significant limitations.
Firstly, the effectiveness of its simple orthogonal constraints on LoRA parameters diminishes as the number of unlearning requests increases, making it difficult to sustainably alleviate cumulative catastrophic utility loss. 
Secondly, using LoRA weights to quantify the degree of unlearning lacks interpretable justification. 
Finally, the mapping from the  Out-Of-Distribution (OOD) detector outputs to the corresponding weights heavily relies on empirical design, which substantially increases the complexity and cost of application.

In this work, we propose Rotation Control Unlearning (RCU), a novel unlearning method that addresses the above challenges. 
This method is inspired by the theory of Lie group (\cite{basics-li}) and re-constructs the LoRA update paradigm through mathematical derivation.
It re-expresses the unlearning update of LLM as rotational operations within a cognitive rotation space.
The cognitive rotation space is defined as a high-dimensional rotation space, which is used to depict the rotational transformations experienced by the original parameters of LLM during continuous unlearning. 
This enables the transformation of the uncontrollable parameter shift into controllable rotational angle changes, thereby effectively alleviating the cumulative catastrophic utility loss.
The specific mathematical formulation is elaborated in the methodology.
Our approach introduces a skew symmetric loss in LoRA update paradigm to formulate the unlearning process as rotation operations, with the rotational angle serving as a precise quantification metric. 
We introduce an orthogonal rotation axes loss to enforce perpendicular rotation directions for consecutive unlearning requests, effectively mitigating cumulative catastrophic utility loss by minimizing inter unlearning request interference.
Furthermore, to enhance compatibility, we design an unlearning alignment loss that guides the OOD detector to produce representations aligned with our LoRA update paradigm.
These representations then collaborate with the distributional shift compensator to generate rotational salience weights for auxiliary quantification.
Finally, our method is supported by straightforward experimental interpretability and requires significantly fewer trainable parameters than $o^3$.

Specifically, our contributions are outlined as follows:
\begin{itemize}
    \item We propose the RCU method, which quantifies the unlearning process by leveraging rotational changes in the cognitive rotation space, and introduce the rotational salience weight to precisely control the degree of unlearning throughout the continuous unlearning process.
    \item We design the skew symmetric loss to establish the existence of the cognitive rotation space and the orthogonal rotation axes loss to alleviate cumulative catastrophic utility loss.
    \item We demonstrate the connection between rotation and unlearning through mathematical proof and experimental validation.
    \item Extensive experiments on the ScienceQA and TOFU datasets confirm the effectiveness of our proposed method without retained dataset.
\end{itemize}

\section{Preliminary}
\textbf{Machine Unlearning in LLMs. }
The objective of machine unlearning is to safeguard information security. 
Currently, there are two mainstream approaches: parameters-based methods~\cite{canshuyouhua1,canshuyouhua2,canshuyouhua3-soul} and in-context unlearning-based methods.~\cite{shangxiawenxiezai1,shangxiawenxiezai2} 
Parameters-based methods iteratively adjust the LLMs' internal parameters to minimize the loss function on specific tasks, thereby improving unlearning performance (\cite{ctrap,exploring,atyaephyra,ails}). 
\cite{ctrap} method employs fine-tuning for rapid learning and induces deliberate model degradation upon detection of harmful fine-tuning behaviors. 
\cite{exploring} approach utilizes a reweighting strategy to adjust training sample weights, focusing particularly on data useful for unlearning. 
In-context unlearning-based methods modify input prompts to prevent the generation of undesired content. 
\cite{unierase} method generates tokens that guide forgetting based on the input query, achieving unlearning without altering model parameters.
Additionally, other techniques exist (\cite{saes,uipe}). 
For instance, \cite{saes}s control forgetting by manipulating model activations. The \cite{5deep} method disrupts the latent space of forgotten samples during training to induce chaotic outputs. 
While existing methods often overlook the challenges of continuous unlearning requests and the associated catastrophic degradation of model utility in real-world scenarios, \cite{o3} formalizes the concept of continuous machine unlearning and introduces an unlearning framework based on an out-of-distribution detector.
Building upon the \cite{o3} paradigm, our method proposes a more refined LoRA update strategy that enables more precise quantification of unlearning extent.

\textbf{Out-Of-Distribution Detection.}
The current methods of OOD detection include one-class SVM based methods~\cite{svmood}, random forest based methods~\cite{suijisenlin}, Gaussian mixture modeling based methods~\cite{gaosimix}, and deep learning based OOD detection methods~\cite{deepood}. 
At the same time, OOD detection based on deep learning has become the mainstream in classification tasks.
Among them, \cite{dagmm} is a method suitable for multi-source time series, which estimates OOD scores by generating low-dimensional representations through deep autoencoders.
\cite{xu-ood} extracts features by pre-trained language model and then fits one-class SVM for detection. 
In addition, \cite{zhou} using ensemble learning, \cite{lang2022estimating} using pseudo-label, \cite{cao} using outlier exposure, \cite{ouyang} using prefix adjustment and other methods have achieved good results in OOD detection.
\cite{o3} incorporates the contrastive entropy loss and Masked Language Modeling  (MLM) loss (\cite{mlm}), enhances the ability to detect out-of-distribution cases.
By studying the unlearning process, we introduced the unlearning alignment loss, thereby enhancing the compatibility between the OOD detector and unlearning.

\begin{figure}[h]
\begin{center}\includegraphics[width=1.01\linewidth]{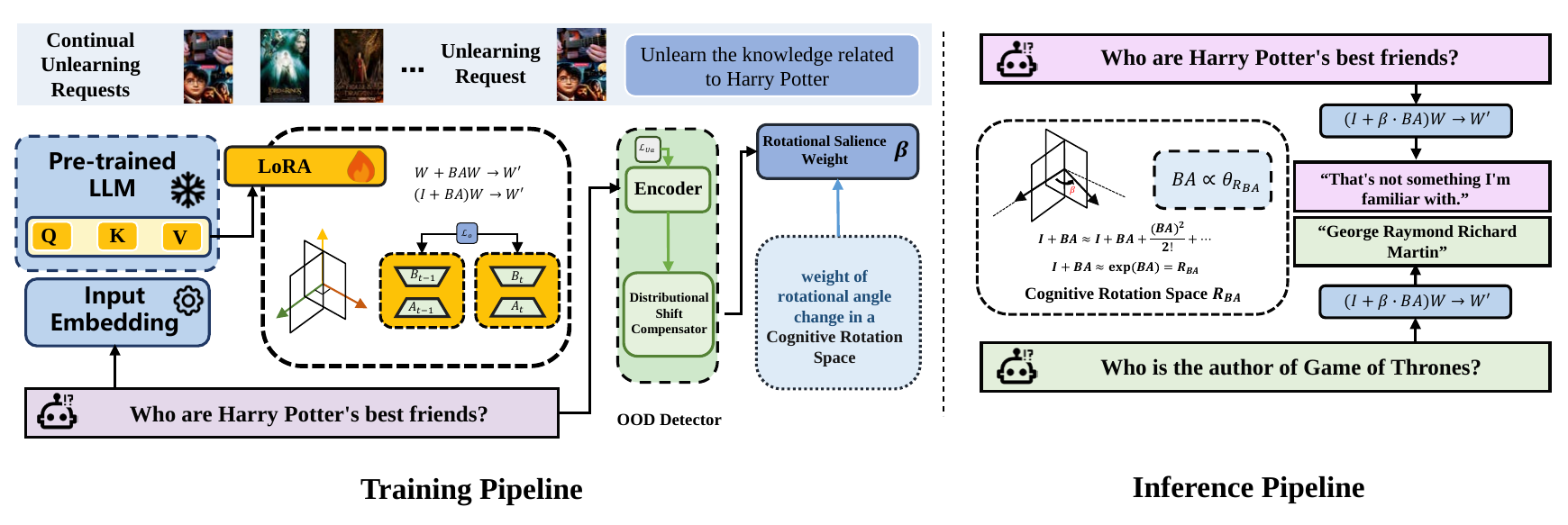}
\end{center}
\caption{The overall architecture of our method is shown in the figure. 
In the training pipeline, the orthogonal rotation axes loss $\mathcal{L} _{o}$ is applied to the attention layers of the LLMs for training; simultaneously, the unlearning alignment loss $\mathcal{L} _{Ua}$ is used to train an OOD detector, whose output is fed into the distributional shift compensator to generate the rotational salience weight $\beta$.
In the inference pipeline, given that the LoRA parameters $BA$ are proportional to the rotation angle $\theta_{R_{BA}}$ in the Cognitive Rotation Space $R_{BA}$.
We control the rotation angle $\theta_{R_{BA}}$ amplitude by adjusting the scale of LoRA $BA$, and use the weight $\beta$ to dynamically load the parameters that match the required unlearning degree.}
\end{figure}


\section{Methodology}

\textbf{Problem Definition.}
We use the popular causal LLMs, where the input to the LLM is a sequence of text tokens of variable length. 
The LLM $M_{\Theta }$, where $\Theta$ is the parameter of the LLM, will calculate the probability of each token in the text under the preorder token based on the input.
We set continuous unlearning ploblem as a series of consecutive arriving unlearning requests, each with $N^{U,t}$ data samples, which can be written as $\left \{ D^{U,t}  \right \} _{t=1}^{T} $.
For the $t-th$ unlearning request , $D^{U,t} = \left \{ x_{i}|| x_{i}\sim \mathcal{P}_{\mathcal{X} }^{U,t}   \right \}  _{i=1 }^{N^{U,t} }$ , where $T$ is the index of the latest arriving unlearning request, and the $P$ is the input marginal distribution.
 In each request, we utilize the input $\mathcal{P}_{\mathcal{X} }^{t}$ and the label distribution $\mathcal{P}_{\mathcal{Y} }^{t}$  for training. 
 Traditional unlearning methods assume a holdout data set drawn from a distribution $\mathcal{P}_{\mathcal{Y} }^{R,t}$ that is disjoint from the forgetting data set $\mathcal{P}_{\mathcal{X} }^{U,t}$ to preserve the performance of the model on the original training distribution. 
 The immediate goal of continuous unlearning is:
\begin{align}
 \sum_{t=1}^{T} \min _{\Theta^{t}} \mathbf{I}\left(M_{\boldsymbol{x} \sim \mathcal{P}_{\mathcal{X}}^{\mathrm{U}, t}}(\boldsymbol{x}, \Theta) ; M_{\boldsymbol{x} \sim \mathcal{P}_{\mathcal{X}}^{\mathrm{U}, t}}^{t}\left(\boldsymbol{x}, \Theta^{t}\right)\right), \sum_{t=1}^{T} \max _{\Theta^{t}} \mathbf{I}\left(M_{\boldsymbol{x} \sim \mathcal{P}_{\mathcal{X}}^{\mathrm{R}, t}}(\boldsymbol{x}, \Theta) ; M_{\boldsymbol{x} \sim \mathcal{P}_{\mathcal{X}}^{\mathrm{R}, t}}^{t}\left(\boldsymbol{x}, \Theta^{t}\right)\right),
  \label{LORAGENGXIN}
\end{align}
where $M^{t}$ with the parameters $\Theta ^{t}$ represents the target model during and after unlearning on the $t-th$ unlearning set $D^{U,t}$, and $I\left ( \cdot ;\cdot  \right ) $ computes the mutual information between two random variables.
The model utility preservation on other distributions $\mathcal{P}_{\mathcal{X} }^{o}$ different from the unlearning distribution is another goal of unlearning.
This can be expressed as follows:
\begin{align}
 \sum_{t=1}^{T} \max _{\Theta^{t}} \mathbf{I}\left(M_{\boldsymbol{x} \sim \mathcal{P}_{\mathcal{X}}^{\mathcal{O}}}(\boldsymbol{x}, \Theta) ; M_{\boldsymbol{x} \sim \mathcal{P}_{\mathcal{X}}^{\mathcal{O}}}^{t}\left(\boldsymbol{x}, \Theta^{t}\right)\right) .
  \label{LORAGENGXIN}
\end{align}

\subsection{Continuous Unlearning for LLM with LoRA}
The continuous unlearning process of LLM inevitably leads to cumulative catastrophic utility loss (\cite{o3}).
The cumulative catastrophic utility loss manifests as a significant decline in both the LLM's unlearning capability and its utility retention capability as the number of unlearning requests increases.
At the parameter level, this is reflected in a shift of the parameters corresponding to previous unlearning requests when the model is trained on new ones.
This requires the method to simultaneously achieve the continuous preservation of its historical unlearning knowledge and the original utility when handling the current unlearning request.


While existing approaches $o^3$ (\cite{o3}) rely on orthogonal constraints to enforce perpendicularity between parameters of $A$ ($W=BA$). 
However, this constraint suffers from inherent limitations: the introduction of parameter $B$ undermines its effectiveness, and its capability further diminishes as the number of unlearning requests accumulates, making it inadequate for continuous unlearning scenarios.
Moreover, the $o^3$ empirically assigns weights to LoRA parameters to represent the degree of unlearning, which is a heuristic design that lacks theoretical grounding and leads to poor interpretability.

We proposed the RCU to address these challenges.
We develop a mathematically-derived approach with enhanced interpretability for quantifying unlearning process.
By constructing a cognitive rotation space wherein LLM parameter updates are formulated as a novel rotational transformation, the RCU transforms continuous unlearning into rotations in a high-dimensional parameter space, thereby converting uncontrollable parameter shift into controlled angular rotations and ultimately alleviating cumulative catastrophic utility loss.

Our analysis in \autoref{result-acc} (a)(b) revealed the relationship between different weighting coefficients $\beta$ and unlearning updates, demonstrating that the extent of unlearning intensifies with increasing values of the $\beta$.

LoRA~\cite{lora} reduces the trainable parameters by introducing two low-rank trainable matrices $ \left \{ A,B \right \}  $, where $W_{LoRA}=BA, W_{LoRA}\in \mathbb{R}^{U\times V}, B\in \mathbb{R}^{U\times K}, A\in \mathbb{R}^{K\times V} $, which decomposes the high-dimensional matrix into a low-rank matrix. 
The specific update formula is as follows:
\begin{align}
 W'\gets W+BA.
  \label{LORAGENGXIN}
\end{align}

LoRA updates inevitably lead to uncontrollable parameter shift.
To address this problem, we draw inspiration from Lie group (\cite{basics-li}) to introduce a new update paradigm. 
This paradigm redefines parameter updates as rotations within the original parameter space. 
As rotations are the rigid transformation (\cite{gangti}), the update via the $BA$ matrices solely governs the change in the rotation angle.
Consequently, the unlearning process is directly driven by changes in this angle, enabling us to use the rotation angle to precisely quantify the degree of unlearning in LLMs.

Firstly, we assume a cognitive rotation space $R$.
Since $R$ can be viewed as an $n$-dimensional rotation matrix, $R$ satisfies the following conditions:
The $R$ is an orthogonal matrix.
The determinant of $R$ is 1, so $det\left ( R \right )=1$.
Since $R$ directly satisfies the two conditions of $\mathit{SO}(n)$, we have $R \in \mathit{SO}(n)$. From~\cite{basics-li}, we can then conclude that cognitive rotation space $R \in \mathit{SO}(n)$ corresponds to at least one matrix $C$ in the Lie algebra $\mathfrak{so}(n)$.

However, every element $C$ in the Lie algebra $\mathfrak{so}\left ( n \right ) $ can be mapped to the $R$ in the Lie group $\mathit{SO} \left ( n \right )$ by the exponential map $exp\left ( C \right ) $. From this we obtain the cognitive rotation space $R$:
\begin{align}
R=exp\left ( C \right ),
  \label{LORAGENGXIN}
\end{align}
here, $C$ is an antisymmetric matrix.
Since $R=exp\left ( C \right ) $, we can obtain from Taylor's Formula:
\begin{align}
R=exp\left ( C \right ) =I+C+\frac{C^2}{2!}+...\approx I+C , C<<I .
  \label{taylor}
\end{align}

From \autoref{taylor}, we can conclude that for any antisymmetric matrix $C$, there exists a corresponding cognitive rotation space $R_{C}$.
Therefore, we construct the skew symmetric loss $\mathcal{L}_{Sk}$ and impose the constraint that $BA$ is an antisymmetric matrix:
\begin{align}
 \mathcal{L}_{Sk}=\left \| \left (BA\right )^T + BA \right \|   _{F}^{2},
  \label{LORAGENGXIN}
\end{align}
here, $I$ is the identity matrix, and $\left \| \cdot  \right \| _{F}^{2} $ is the Frobenius norm.

In addition, due to the influence of factors such as the learning rate, $BA << I$ (as summarized in \autoref{parm} and \autoref{LORAba} of Appendix A.6, the specifications for the $BA$ matrix parameters are detailed there.).
Thereby, there exists a cognitive rotation space $R_{BA}\approx I+BA$.

We can establish the following LoRA update paradigm:
\begin{align}
 W\gets W+BAW=\left ( I+BA \right ) W,
  \label{gengxin}
\end{align}
here, we update the above equation given a set of low-rank parameters $ \left \{ A,B \right \}  $, where the parameter matrix $W$ of LLM is frozen. 

From \autoref{gengxin}, we can consider the update of the $W$ as a rotation in the cognitive rotation space $R_{BA}$.


After expressing the update of $BA$ as the cognitive rotation space $R_{BA}$ in the parameter space, we aim to quantify the degree of unlearning of LLM by changing the rotation angle of $R_{BA}$.

Although we only trained the attention layers, due to the huge parameter size of the large language model, directly calculating the rotation angle $\theta$ of $R_{BA}$ still incurs a significant amount of computational cost.
Here we present our theorem 1.

\textbf{Theorem 1:} \textbf{\textit{For $R\in \mathbb{R} ^{n\times n} $, when $R=exp\left ( C \right ) $, the rotation angle $\theta$ of $R$ is directly proportional to $C$.}}
The proof of theorem 1  can be found in the appendix.

From \textbf{Theorem 1}, if we want to change the rotational angle $\theta $ of $R_{BA}$, we only need to make the corresponding changes to $BA$.
Therefore, we obtain the rotational salience weight $\beta$ from the OOD detector and the distributional shift compensator.
When $\beta \times BA$ , all the rotation angles $\theta $ in $R_{BA}$ are changed to $\beta \times \theta$.
Therefore, we can use the rotation Angle $\theta $ to quantify the unlearning degree of LoRA, and only need to use $BA$ for the calculation.



In addition, for achieving effective unlearning, we utilize preference optimization to update the model to accommodate random task labels or refuse-based answers such as "I don't know", which we call $y'$. 
For each unlearning we only train the cross-entropy loss using the unlearning dataset of our current knowledge:
\begin{align}
\mathcal{L}_{CE} =-\frac{1}{N^{U,t} }\sum_{i=1}^{N^{U,t} }{y'}_{i}^{U,t}\log{M_{\Theta }\left ( x_{i}^{U,t}   \right )  } .
  \label{LORAGENGXIN}
\end{align}

To reduce the interaction between the change of the rotation angle for each unlearning request in continuous unlearning, we make the rotation axes of each rotation perpendicular to each other.
The rotation axes here refers to the subspace formed by the points that remain stationary under rotation in the high-dimensional space.

Here, we know that when unlearning request $t$, the corresponding cognitive rotation space is $R_{B_{t}A_{t}}$.
Then, from request $t-1$ to request $t$, the relative rotation matrix is $\bigtriangleup R_{t}$. 
From this, we can obtain :
\begin{align}
\bigtriangleup R_{t}=R_{t}\cdot R_{t-1}^{T} 
=\left ( I+B_{t}A_{t} \right ) \left ( I+B_{t-1}A_{t-1} \right ) ^{T} ,
  \label{LORAGENGXIN}
\end{align}
since $BA$ is an antisymmetric matrix, we have:
\begin{align}
\bigtriangleup R_{t}&=\left ( I+B_{t}A_{t} \right ) \left ( I+B_{t-1}A_{t-1} \right ) ^{T}  
=\left ( I+B_{t}A_{t} \right ) \left ( I-B_{t-1}A_{t-1} \right ) \\ \nonumber
&=I+B_{t}A_{t}-B_{t-1}A_{t-1}-B_{t}A_{t}B_{t-1}A_{t-1} 
\approx I+B_{t}A_{t}-B_{t-1}A_{t-1},
\label{LORAGENGXIN444}
\end{align}

We cannot directly calculate the rotation axes for the calculation because this would consume a large amount of computing resources and significantly slow down the training speed of the model. 
Here, we know from \textbf{Theorem 2} that when the cognitive rotation space $R_{B_{t-1}A_{t-1}}$, $R_{B_{t}A_{t}}$ and $B_{t-1}A_{t-1}$, $B_{t}A_{t}$ are mutually perpendicular, their rotation axes must be perpendicular.
Therefore, we make the cognitive rotation space $R_{B_{t-1}A_{t-1}}$ of the $(t-1)$-th request and the relative rotation space $\bigtriangleup R$ relative to the $t$-th request and the $(t-1)$-th request mutually perpendicular. 
This ensures that the rotation angles of each unlearning request in the cognitive rotation space do not affect each other, reducing the cumulative catastrophic utility loss generated with continuous unlearning.

\textbf{Theorem 2:} 
when $R=exp\left ( A \right )$ and $R'=exp\left ( A' \right )$ , $A\bot A'$, then the rotation axes of $R$ and $R'$ are perpendicular to each other.
The proof of theorem 2  can be found in the appendix.

We hope that $\bigtriangleup R_{t}=I+B_{t}A_{t}-B_{t-1}A_{t-1}$ and $R_{t-1}=I+B_{t-1}A_{t-1}$ are perpendicular to each other, then $B_{t}A_{t}-B_{t-1}A_{t-1}$ and $B_{t-1}A_{t-1}$ will also be perpendicular to each other from \autoref{taylor} and Theorem 2.
The orthogonal rotation axes loss are as follows:
\begin{align}
\mathcal{L}_{o}=\left \| \left ( W_{t} -W_{t-1}  \right ) \cdot W_{t-1} \right \|  _{F}^{2} =\left \| \left (  B_{t}A_{t} -B_{t-1}A_{t-1} \right ) \cdot \left ( B_{t-1}A_{t-1} \right )  \right \|  _{F}^{2} ,
\end{align}
where $W_{t-1}=B_{t-1}A_{t-1}$ are the parameters of the lora after training on the $\left (t-1\right )$-th request.
The $\left \| \cdot  \right \|  _{F}^{2}$ is the Frobenius norm.

In summary, the overall loss of our method is as follows:
\begin{align}
\mathcal{L}_{overall} =\lambda _{1}  \mathcal{L}_{Sk}  +\lambda _{2}\mathcal{L}_{o}+\lambda _{3}\mathcal{L}_{CE} ,
  \label{44444}
\end{align}
here,we set $\lambda _{1} =0.1$,$\lambda _{2} =0.1$ and $\lambda _{3} =1$ on the ScienceQA dataset.
We set $\lambda _{1} =0.01$,$\lambda _{2} =0.5$ and $\lambda _{3} =1$ on the TOFU dataset.

\subsection{Unlearned Knowledge Detection}

\textbf{OOD Detection.}
Based on $o^3$, we turn the unlearned knowledge detection task into an OOD task by treating the unlearned dataset as In-Distribution (ID) data, and leverage a scoring mechanism to quantify the extent of unlearning.

We propose the OOD detection loss, which consists of three parts.
We use the contrastive entropy loss and Masked Language Modeling (MLM) loss (\cite{o3}).
As shown in \autoref{result-acc}, the updates of RCU exhibit an uneven characteristic, where the feature always involves continuous updates within a very small range of rotation angle changes. 
Given this characteristic, in order to make the output of OOD detection better align with the update pattern of RCU, we introduced the unlearning alignment loss $\mathcal{L} _{Ua}$.
The $\mathcal{L} _{Ua}$ are as follows:
\begin{align}
\mathcal{L}_{Ua}=\frac{1}{d^{2} }   \left \| \frac{\hat{Z}_{i}^{T} \hat{Z}_{i}}{n-1} -I  \right \|_{F}^{2},  
 \text { where } \hat{Z}_{i}=\frac{Z_i}{\left \| Z_i \right \|_{2}  } ,
  \label{LORAGENGXIN}
\end{align}
where $\left \| \cdot  \right \| _{2} $ is $L_2$ norm.
The $Z_i$ is the average pooled feature representation from layer i of the backbone network.
The $\left \| \cdot  \right \| _{F}^{2} $ denotes the Frobenius norm.


In addition, the contrastive entropy $\mathcal{L} _{CEL} $ also starts with the augmentation view generation.
The $\mathcal{L} _{CEL}$ (\cite{o3}) leverage random masking to generate the ﬁrst view type. 
For a particular text instance $x$ with tokens of length $n$, we randomly select $p\%$ ($p = 15$ in our implementation) tokens and replace them with the tokens of [MASK].
The $x^{\ast }$ is the instance with the random masking.
For the second contrastive view, we make use of a key encoder $F_{\Omega ^{key} }$, which is initialized from the original OOD module backbone $F_{\Omega}$ that is a transformer consisting of $L$ attention layers : $F:=f_{\omega_{1}} \circ \cdots f_{\omega_{l}} \circ \cdots f_{\omega_{L}}$.
Then we input the original text instance $x$ and generate the second view from $F_{\Omega ^{key} }$.
The $\mathcal{L} _{CEL}$ is as follow:
\begin{align}
\mathcal{L}_{\mathrm{CEL}}=-\sum_{i=1}^{N^{B}} \sum_{l=1}^{L} \sum_{j=1}^{N^{B}} \Delta(i, l, j) \log (\Delta(i, l, j)),  \\ \nonumber
 \textbf{where }  \Delta(i, l, j)=\frac{\exp \left(f_{\omega_{[1: l]}}\left(\boldsymbol{x}_{i}^{*}\right) \cdot f_{\omega_{[1: l]}^{\mathrm{key}}}\left(\boldsymbol{x}_{j}\right)\right)}{\sum_{k=1}^{N^{B}} \exp \left(f_{\omega_{[1: l]}}\left(\boldsymbol{x}_{i}^{*}\right) \cdot f_{\omega_{[1: l]}^{\mathrm{key}}}\left(\boldsymbol{x}_{k}\right)\right)},
  \label{LORAGENGXIN}
\end{align}
here $N$ is the sample quantity of a mini-batch. And the $f_{\omega_{[1: l]}}\left(\boldsymbol{x}_{i}^{*}\right)$ is the token averaging representation of the $l$-th layer. 
We use MLM loss $\mathcal{L} _{MLM}$ (\cite{mlm}) to improve the language generation of our model:
\begin{align}
\mathcal{L}_{MLM}=-\frac{1}{N^{B} }   \sum_{i=1}^{N^{B} }y_{i}^{\ast }  \log{F_{\Omega }\left ( x_{i}^{\ast } \right )  },
\end{align}
where $y^{\ast }$ is the random token masking label.
Here, the $\mathcal{L} _{CEL}$ focuses on the relative relationship of sample pairs. The $\mathcal{L} _{MLM}$ boosts the representation power of the generated language.

The final loss $\mathcal{L} _{OOD}$ can be:
\begin{align}
\mathcal{L} _{OOD} = \mathcal{L} _{CEL}+ \mathcal{L} _{MLM} +\mathcal{L}_{Ua} .
  \label{LORAGENGXIN}
\end{align}

\textbf{Distributional Shift Compensator.}
We follow the method for obtaining the output of OOD detection as described in $o^3$. 
We utilized the Mahalanobis distance and the distance based on the maximum instance cosine similarity. 
Finally, we calculated the combined score $\gamma^t$.
For the calculation of the OOD score $\gamma^t$, please refer to the appendix.

After we get combined score $\gamma^t$, we need to map the $\gamma^t$ into an rotational salience weight $\beta$. 
Here, we hope that the change of $\beta$ can conform to the unlearning process of the cognitive rotation space $R_{BA}$.
However, as the unlearning learning proceeds, we find that the performance of the update based on RCU is uneven. 
As shown in the \autoref{result-acc} (a)(b), the model does not learn unlearning knowledge before $\beta=0.3$  on the ScienceQA dataset. 
At $\beta=0.3$ to $\beta=0.5$, the unlearned knowledge is gradually learned, while at $\beta=0.5$ to $\beta=1$, the model has fully learned the unlearned knowledge.
On the TOFU dataset, the range of knowledge that the model learns for unlearning is approximately between $\beta=0.2$ and $\beta=0.6$. 
The specific results can be found in the appendix.
This gives us the following relation:
\begin{align}
\beta =\left\{\begin{matrix}
 0.45
 & \Gamma _{2}< \gamma^{t}\le 1  ,\\
 \mathcal{M}\left ( \gamma^{t}  \right )   & \Gamma _{1} < \gamma^{t} \le \Gamma _{2} ,\\
 0 & \gamma^{t} \le \Gamma _{1} ,
\end{matrix}\right.
  \label{yingshe}
\end{align}
the $\Gamma _{1}$ and $\Gamma _{3}$ are thresholds, which $\Gamma _{1}=1e-80$, $\Gamma _{2}=0.1$ on the ScienceQA dataset and $\Gamma _{1}=0.2$, $\Gamma _{2}=1$ on the TOFU dataset.
The $\mathcal{M}\left ( \gamma^t  \right ) $ on the ScienceQA dataset is $0.35+\left ( \log_{10}{\gamma^t } +80 \right )/790$.
The $\mathcal{M}\left ( \gamma^t  \right ) $ on the TOFU dataset is $0.35 + \left ( \left ( \gamma^t  - 0.2 \right )  / 0.8 \right )  \cdot  0.25$.

Finally, for each input $x$ of the $t$-th unlearning request, the corresponding parameters $W_{x}^{t}$ can be expressed as:
\begin{align}
W_{x}^{t}=\left ( I+\beta \cdot BA \right )W .
  \label{yingshe}
\end{align}

\section{Experiments}

\subsection{Datasets}
We conducted experiments on two tasks: question answering and fictional knowledge generation. 
We have divided the question-answering task into 5 consecutive sub-tasks, and the fictional knowledge generation has been divided into 3 consecutive sub-tasks.
The detailed introduction of the datasets is as follows:

\textbf{Question Answeing.}
We use ScienceQA (\cite{scienceqa}) as the question and answer dataset. 
This dataset consists of 6,508 training samples and 2,224 testing samples.
We selected five of these areas as the continuous unlearning requests, namely biology, physics, chemistry, economics, and earth-science.
We utilized the CommonsenseQA (\cite{commonsenseqa}) as the utility dataset, which contained 9,740 training samples and 1,221 validation samples, to evaluate the commonsense reasoning ability of LLMs. 
The OpenbookQA (\cite{openbookqa}) can assess the understanding ability of books.
The training set contains 4,957 samples, the validation set includes 500 samples, and the test set consists of 500 samples.

\textbf{Fictitious Knowledge Generation.}
We conducted a test on the generation of fictional knowledge using the TOFU dataset (\cite{tofu}).
The TOFU dataset contains questions about fictional authors synthesized by GPT-4. 
The three unlearning sets 'foget01', 'foget05', and 'foget10' respectively represent random selection ratios of 1\%, 5\%, and 10\% of the authors. 
The authors in each unlearning set are mutually exclusive.
Additionally, we also utilized the data related to real-word authors and world facts in this dataset to test the LLMs' ability to maintain its effectiveness.

\subsection{Experimental Setup}

\begin{table*}[h]
 \caption{Performance Comparison between our method and other baselines when continually unlearning TOFU-forget01, -forget05, and -forget10 in Fictitious Knowledge Generation. 
 The * represents the results we achieved in our own experimental environment.
 }

\renewcommand\arraystretch{1}
  \centering
  \resizebox{1\columnwidth}{!}{
  \begin{tabular}{c|ccccc|ccccc|ccccc|}
    \toprule
      Method & \multicolumn{5}{c}{Unlearning Request 1}& \multicolumn{5}{c}{Unlearning Request 2}& \multicolumn{5}{c}{Unlearning Request 3}\\
          &  S.U.$\downarrow$ & D.U.$\downarrow$ & R.D.$\uparrow$ & R.A.$\uparrow$ & W.F.$\uparrow$ &  S.U.$\downarrow$ & D.U.$\downarrow$ & R.D.$\uparrow$ & R.A.$\uparrow$ & W.F.$\uparrow$ &  S.U.$\downarrow$ & D.U.$\downarrow$ & R.D.$\uparrow$ & R.A.$\uparrow$ & W.F.$\uparrow$ \\
   
    \midrule
    Base *  &85.0&90.0&85.8&89.0&87.0 &87.3 &89.3&85.8&89.0&87.0&85.3& 90.0 &85.8&89.0&87.0 \\
    \midrule
    GradASC  &75.0&85.0&71.0&86.0&82.1&17.6&23.1&19.0& 0&0&17.1&\textbf{14.2}&19.0&0&0 \\
    GradDif  &78.1&84.0&81.9&86.7&83.5&62.5&70.0&70.4&65.7&77.9&\textbf{16.5}&15.2&19.0&0&0 \\
    EUL  &84.1&86.3&\textbf{86.1}&86.7&87.1&84.4&90.3&\textbf{85.8}&88.0&85.5&80.1&83.5&83.4&86.3&83.5 \\
    PO * &18.75&25.0&77.0&85.0&81.0&31.88&52.5&79.0&86.0&79.0&43.13&52.5&77.75&78.0&80.0 \\
    NPO  &68.8&75.0&83.6&89.0&81.8&76.3&84.2&83.2&87.7&84.1&77.6&79.2&81.4&87.3&82.9 \\
    SOGD  &43.7&76.0&80.3&85.3&83.4&22.8&24.0&79.0&81.3&82.6&17.4&21.7&82.3&77.0&82.1 \\
    SOPO * &31.25&37.5&83.7&85.0&83.0&38.13&45.0&80.0&87.0&82.0&36.88&43.75&79.5&85.0&82.0 \\
    $o^3$ *  &15.66&14.67&85.25&89.0&86.3&22.49&20.17&85.5&89.0&86.3&26.66&23.56&85.25&89.0&86.3 \\
    \midrule
    \rowcolor{gray!20}
    Ours  &\textbf{9.37}&\textbf{12.5}&85.57&\textbf{89.0}&\textbf{87.0}&\textbf{12.5}&\textbf{12.5}&85.61&\textbf{89.0}&\textbf{87.0}& 20.6&17.5&\textbf{85.6}&\textbf{89.0}&\textbf{87.0} \\
        
    \bottomrule
  \end{tabular}
  }
  \label{TOFU-RE}
\end{table*}

\textbf{Evaluation Metrics.}
We continue to use the $o^3$ (\cite{o3}) key evaluation indicators. 
Here, the Sample-level Unlearning (S.U.) represents the performance of the test LLMs on the unlearning training set when there is a unlearning request.
The Distribution-level Unlearning (D.U.) indicates the performance of the test LLMs on the unlearning test set when there is the unlearning request. 
In addition, we use three indicators to measure the performance of the LLMs in maintaining utility. 
The Retained Distribution (R.D.) represents the distribution that is most sensitive to unlearning requests.
The CommonsenseQA (C.QA.) and QpenbookQA (O.QA.) are datasets used on the scienceQA dataset to measure the utility of each request.
We use the accuracy of these two datasets to measure their performance in QA.
On the TOFU dataset, we use the utility datasets provided by the two datasets, namely Real-word Authors (R.A.) and Word Fact (W.F.), to measure their performance in Fictitious Knowledge Generation.

\textbf{Compared Baseline.}
We compared a series of the most advanced LLM unlearning methods: GradAsc~\cite{gradasc}, GradDif~\cite{graddif}, EUL~\cite{EUL}, PO~\cite{canshuyouhua2}, NPO~\cite{NPO}, SOGD~\cite{canshuyouhua3-soul}, SOPO~\cite{canshuyouhua3-soul} and $o^3$~\cite{o3}.
The base refers to the result obtained directly through the LLM testing.

\textbf{Implementation Details.}
We use LLaMA2-7b (\cite{llama7B-2}) as the target model. 
The detection backbone is Pseudo-Roberta-Large (\cite{roberta}). 
For the TOFU dataset, the learning rate is 2e-4, and the number of epoch is 10. 
For the ScienceQA dataset, the batch size is 128, the number of epoch is 15, and the learning rate is 3e-4. 
In the inference pipeline of ScienceQA, the max batch size is 24. 
The LoRA ranks for both datasets are 8.
Our method only fine-tunes the attention layers in LLM.
We conducted our experiments on two NVIDIA RTX A6000.

\subsection{Experimental Results}

\textbf{Question Answering.}
The results of our method are shown in appendix A.8 and \autoref{scqa-acc}.
We compared base, PO, SOPO, $o^3$ and our results. 
Some detailed results can be found in the appendix.
Our method achieved the same results as the base method (C.QA. and O.QA.). 
The results of the R.D. were also very similar to those of the base method.
However, our method performed much better than the others in the S.U. and D.U. indicators.
Compared with the most advanced method $o^3$, the S.U. decreased by 9.68\%, 4.02\%, 4.05\%, 7.58\% and 10.82\% respectively on five requests.
The D.U. compared to the most advanced method $o^3$, on average, decreased by 17.67\% across five requests.
In addition, the number of training parameters of our method is much smaller than that of $o^3$. 

   
        
  

\textbf{Fictitious Knowledge Generation. }
\autoref{TOFU-RE} presents the experimental results of our method on the TOFU dataset. 
Based on the results, we found that our method can effectively enhance the LLMs' ability to unlearning (D.U. and S.U.).
Compared with the currently best continuous unlearning method $o^3$, the result on the S.U. decreased by 6.29\%, 9.99\% and 6.06\%.
the result on the D.U. decreased by 2.17\%, 7.67\% and 6.06\%.
Furthermore, in terms of the stability of the LLMs' utility, the performance level of our method is comparable to the current best results.
We observed that as the number of unlearning requests increased, the unlearning accuracy on the test set (D.U.) of most methods showed significant fluctuations or increasing.
This phenomenon indicates that catastrophic utility loss continues to accumulate, and the LLMs' stability is affected.
In contrast, our method not only effectively mitigates such utility losses but also significantly improves the performance stability of LLMs during the continuous unlearning process.

In two datasets, $o^3$ has 19.99 M trainable parameters, compared to our method's 8.39 M.
The required number of training parameters is significantly less than that of $o^3$.
The number of parameters that need to be updated in our method is much smaller than that in $o^3$.

\begin{figure*}[t]
\centering
\subfloat[]{\includegraphics[width=1.8in]{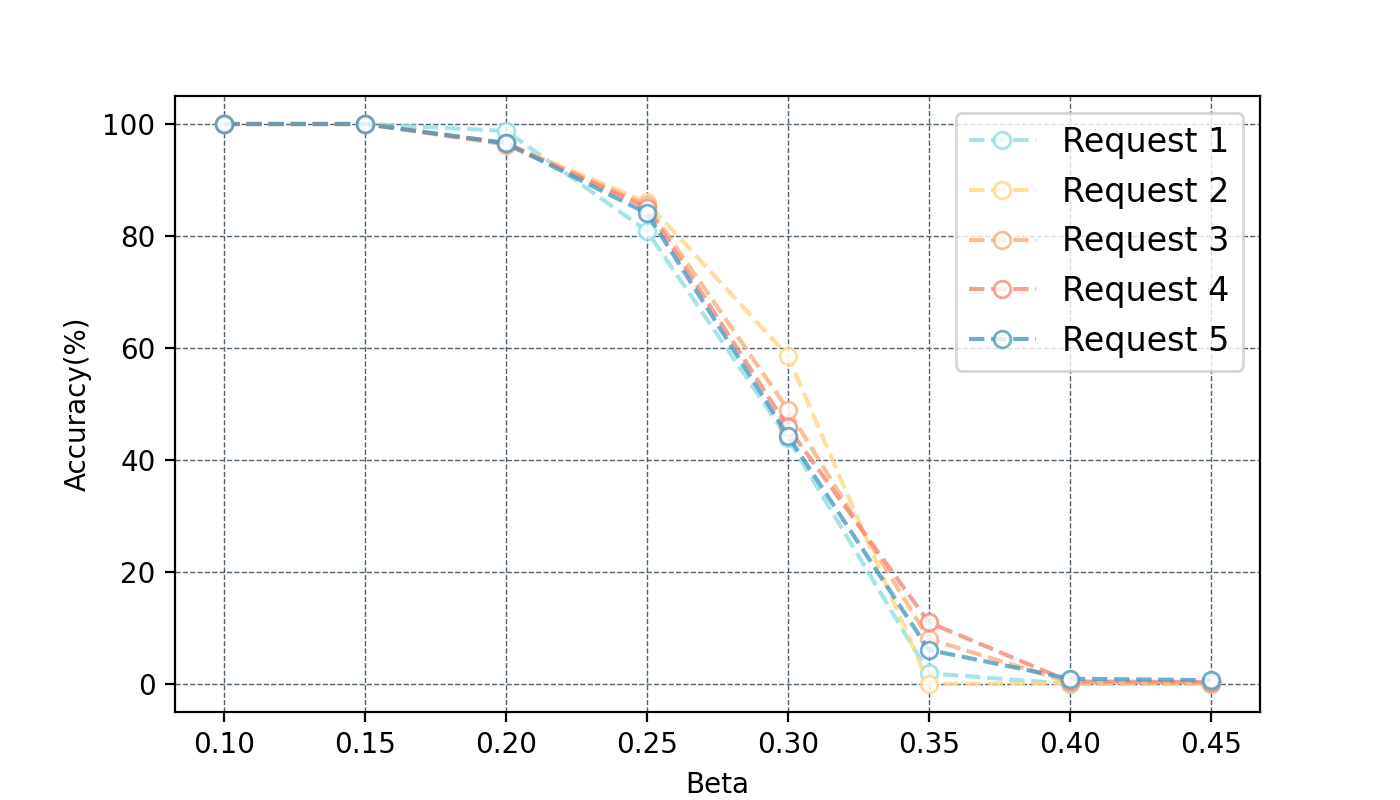}%
   \label{imagenetr-1}}
\hfil
\subfloat[]{\includegraphics[width=1.8in]{ 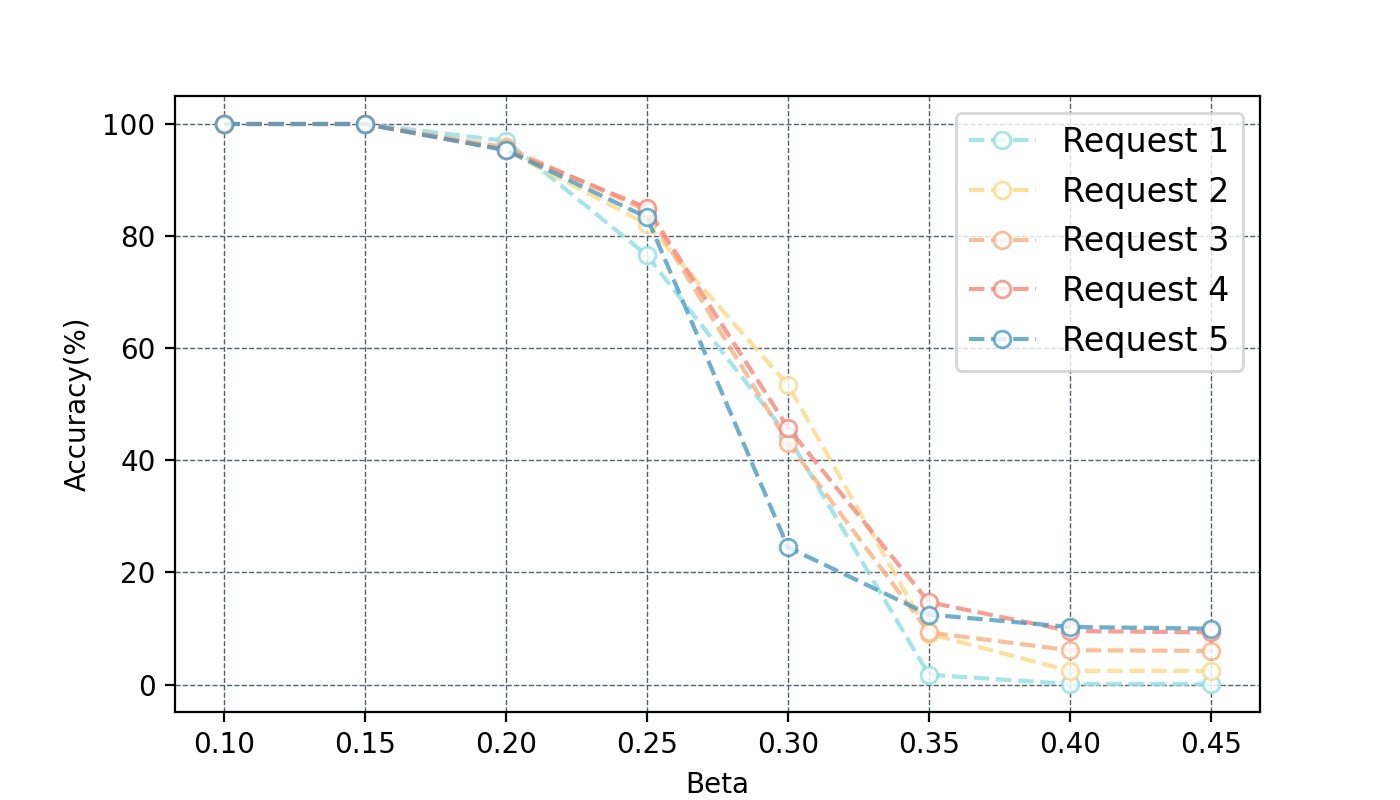}%
   \label{imagenetr-2}}
   \hfil
\subfloat[]{\includegraphics[width=1.8in]{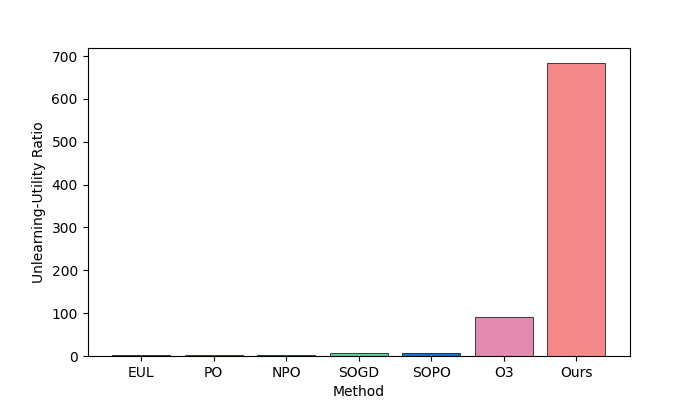}
   \label{imagenetr-2}}
\hfil
 \caption{Experimental results on the ScienceQA dataset.
 (a) The relationship between the $\beta$ ($\beta$ is proportional to the rotation angle) and unlearning processes (S.U.).
    (b) The relationship between the $\beta$ ($\beta$ is proportional to the rotation angle) and unlearning processes (D.U.).
    (c) The results of $U^2R$ on the TOFU dataset.
    }
\label{result-acc}
\end{figure*}

Furthermore, we follow the Unlearning-Utility Ratio ($U^2R$) in $o^3$.

\begin{align}
U^2R = \frac{\mathrm{Acc}_{\mathrm{S} . \mathrm{U} .}^{0}+\mathrm{Acc}_{\mathrm{D} . \mathrm{U} .}^{0}-\mathrm{Acc}_{\mathrm{S} . \mathrm{U} .}^{T}-\mathrm{Acc}_{\mathrm{D} . \mathrm{U} .}^{T}}{\mathrm{Acc}_{\mathrm{R} . \mathrm{D} .}^{0}+\mathrm{Acc}_{\mathrm{U} .1 .}^{0}+\mathrm{Acc}_{\mathrm{U} .2 .}^{0}-\mathrm{Acc}_{\mathrm{R} . \mathrm{D} .}^{T}-\mathrm{Acc}_{\mathrm{U} .1 .}^{T}-\mathrm{Acc}_{\mathrm{U} .2 .}^{T}}  ,
  \label{LORAGENGXIN}
\end{align}
where the Acc denotes Accuracy, the U.1 represents C.QA or R.A., the U.2 represents O.QA or W.F., and the T denotes the unlearning request.
The result on the TOFU dataset is shown in \autoref{result-acc} (c).
Our method has demonstrated significant advantages in both unlearning and utility retention. 
Furthermore, the results on the ScienceQA dataset can be found in the Appendix A.2.

\subsection{The Relationship between Rotation and Unlearning}
In \autoref{result-acc} (a) and \autoref{result-acc} (b), we have presented the changes about $\beta$, along with the experimental results for each unlearning request. 
Based on the results on the ScienceQA dataset, we can observe that when $\beta$ is within the range of 0.15 to 0.45, the model rapidly undergoes the process of unlearning. 
However, when $\beta$ is less than 0.15, the model does not forget the knowledge. 
And when $\beta$ is greater than 0.45, the model has achieved complete forgetting and no longer continues to unlearning.
The above experimental results indicate that the unlearning process of RCU is achieved within a very small rotation angle change range in the cognitive rotation space.
Therefore, when designing the OOD detector and the distributional shift compensator, we must also ensure that the rotational salience weights they output have a relatively concentrated distribution to match the above characteristics.
The $\beta$ changes regarding TOFU dataset are presented in the appendix.



\subsection{Ablation Study}
The results of our ablation experiments are shown in \autoref{AB}. 
Here, we aim to focus more on the unlearning process rather than the OOD process. 
Therefore, we consider the impact of $\mathcal{L}_{Ua}$ on the unlearning performance rather than on the OOD performance. 
Based on the results, we found that $\mathcal{L}_{Sk}$ is the main influencing factor for the unlearning performance.
And the LoRA update Paradigm (RC-LoRA in \autoref{AB}), $\mathcal{L}_{Sk}$ and the $\mathcal{L}_{o}$ are three key factors for mitigating cumulative catastrophic utility loss in continuous unlearning.
The results of the ablation experiment proved the effectiveness of our method.

\begin{table}[htbp]
 \caption{The ablation experiment results on the ScienceQA dataset.The RC-LoRA represents our LoRA update paradigm.}
\renewcommand\arraystretch{1}
  \centering
  \resizebox{1\columnwidth}{!}{
  \begin{tabular}{c|ccccc|ccccc|ccccc|l}
    \toprule
      Method & \multicolumn{5}{c}{Unlearning Request 1}& \multicolumn{5}{c}{Unlearning Request 2}& \multicolumn{5}{c}{Unlearning Request 3} \\
          &  S.U.$\downarrow$ & D.U.$\downarrow$ & R.D.$\uparrow$ & C.QA.$\uparrow$ & O.QA.$\uparrow$ &  S.U.$\downarrow$ & D.U.$\downarrow$ & R.D.$\uparrow$ & C.QA.$\uparrow$ & O.QA.$\uparrow$  &  S.U.$\downarrow$ & D.U.$\downarrow$ & R.D.$\uparrow$ & C.QA.$\uparrow$ & O.QA.$\uparrow$  \\
    \midrule
    w/o $\mathcal{L}_{o}$ &0.54&0.68&89.01&79.19&83.09&0.65&4.74&88.02&79.19&83.09&0.78&8.98&88.22&79.19&83.09\\
    w/o $\mathcal{L}_{Sk}$  &99.75&98.99&90.16&79.19&83.09&99.51&98.01&89.18&79.19&83.09&99.51&98.04&88.44&79.19&83.09 \\
    w/o RC-LoRA
    &0.12&1.35&90.51&79.19&82.39&0.58&4.24&91.11&79.19&82.39&0.99&22.45&90.68&79.19&82.39\\
    w/o $\mathcal{L}_{Sk}$+RC-LoRA 
    &100.00&99.24&89.16&79.19&83.09&99.52&98.14&88.25&79.19&83.09&99.62&98.53&87.44&79.19&83.09\\
     w/o $\mathcal{L}_{Ua}$ 
     &1.27&3.44&87.01&79.02&81.35&2.77&4.85&88.08&79.02&81.35&3.11&8.23&87.62&79.02&81.35\\
    \midrule
    \rowcolor{gray!20}
    Ours  &\textbf{0.05}&\textbf{0.12}&\textbf{92.5}&\textbf{79.19}&\textbf{83.2}&\textbf{0.17}&\textbf{2.45}&\textbf{91.75}&\textbf{79.19}&\textbf{83.2}&\textbf{0.33}&\textbf{6.13}&\textbf{90.5}&\textbf{79.19}&\textbf{83.2} \\
        
    \bottomrule
  \end{tabular}
  }

\renewcommand\arraystretch{1}
  \centering
  \resizebox{0.8\columnwidth}{!}{
  \begin{tabular}{c|ccccc|ccccc|}
    \toprule
      Method & \multicolumn{5}{c}{Unlearning Request 4}& \multicolumn{5}{c}{Unlearning Request 5}\\
           &  S.U.$\downarrow$ & D.U.$\downarrow$ & R.D.$\uparrow$ & C.QA.$\uparrow$ & O.QA.$\uparrow$ &  S.U.$\downarrow$ & D.U.$\downarrow$ & R.D.$\uparrow$& C.QA.$\uparrow$ & O.QA.$\uparrow$  \\
    \midrule
     w/o $\mathcal{L}_{o}$ &1.24&11.58&88.06&79.19&83.09&2.41&16.85&87.95&79.19&83.09\\
     w/o $\mathcal{L}_{Sk}$  &99.75&98.04&88.20&79.19&83.09&99.75&98.04&87.95&79.19&83.09 \\
     w/o RC-LoRA
     &1.03&30.68&89.43&79.19&83.09&2.46&31.85&\textbf{90.00}&79.19&83.09 \\
     w/o $\mathcal{L}_{Sk}$+RC-LoRA 
     &99.85&98.53&87.2&79.19&83.09&99.85&98.28&86.45&79.19&83.09 \\
      w/o $\mathcal{L}_{Ua}$ 
      &9.19&11.07&87.33&79.02&81.35&13.65&15.21&86.93&79.02&81.35\\
    \midrule
    \rowcolor{gray!20}
    Ours  &\textbf{0.49}&\textbf{9.56}&\textbf{90.5}&\textbf{79.19}&\textbf{83.2}&\textbf{0.98}&\textbf{10.29}&89.753&\textbf{79.19}&\textbf{83.2}\\
        
    \bottomrule
  \end{tabular}
  }
  \label{AB}
\end{table}



\begin{figure*}[!t]
\centering
\subfloat[]{\includegraphics[width=1.8in]{ 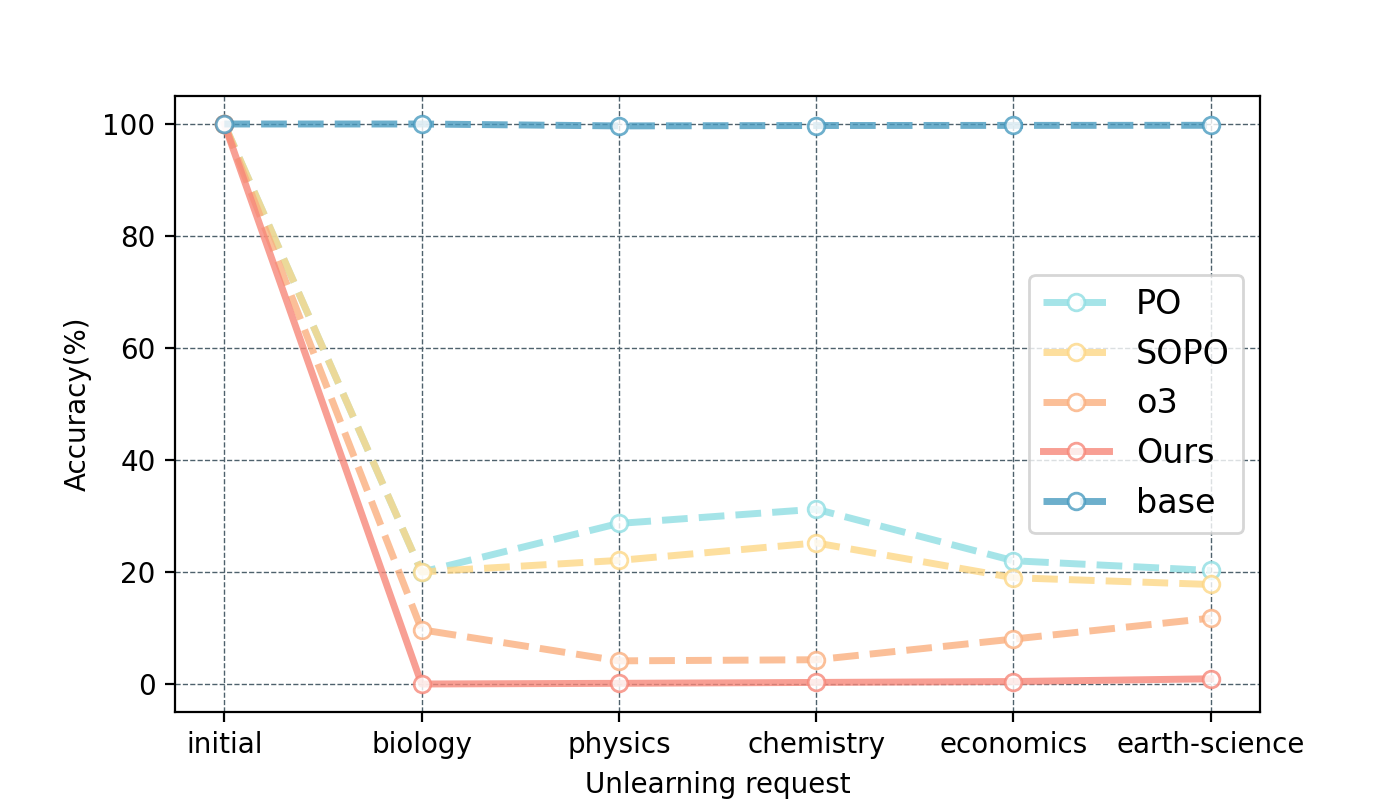}%
   \label{imagenetr-1}}
\hfil
\subfloat[]{\includegraphics[width=1.8in]{ 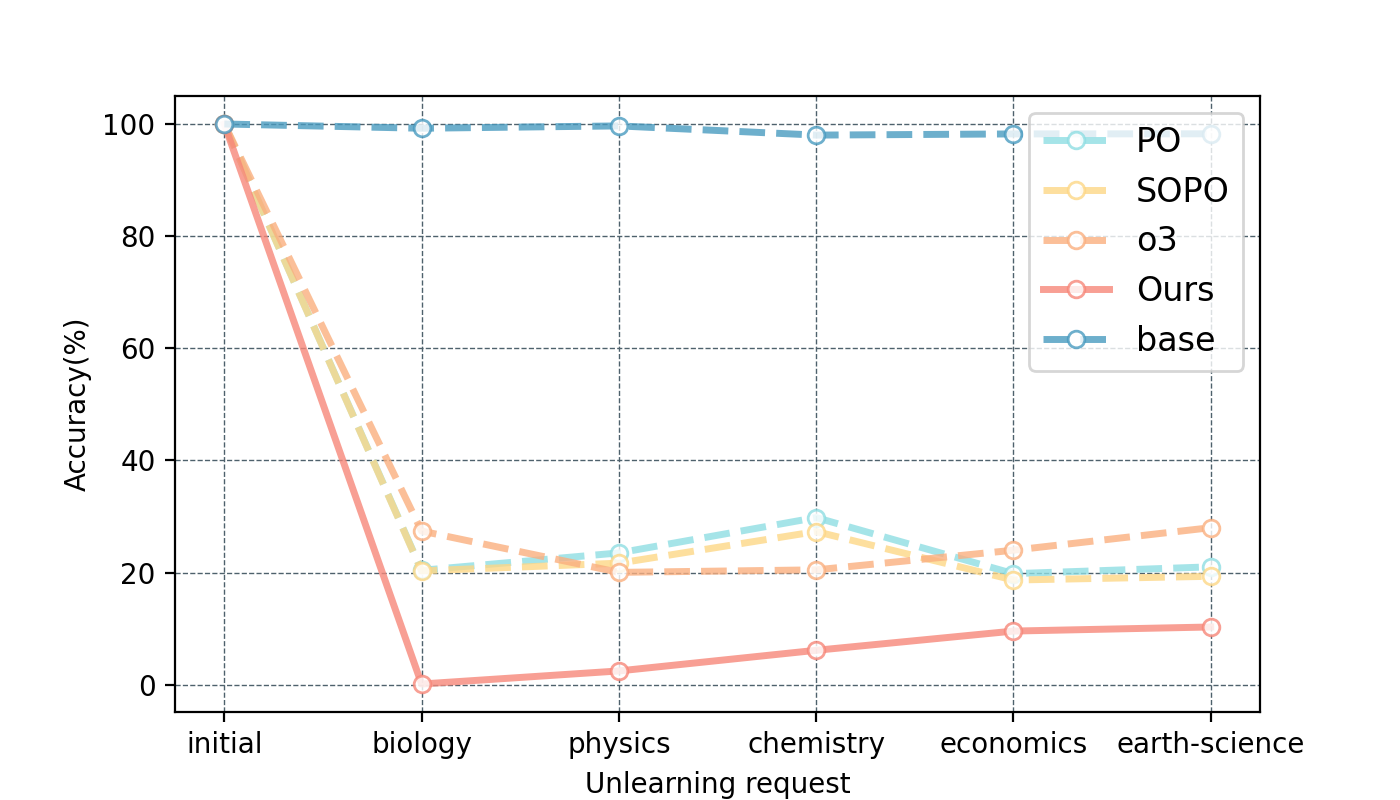}%
   \label{imagenetr-2}}
\hfil
\subfloat[]{\includegraphics[width=1.8in]{ 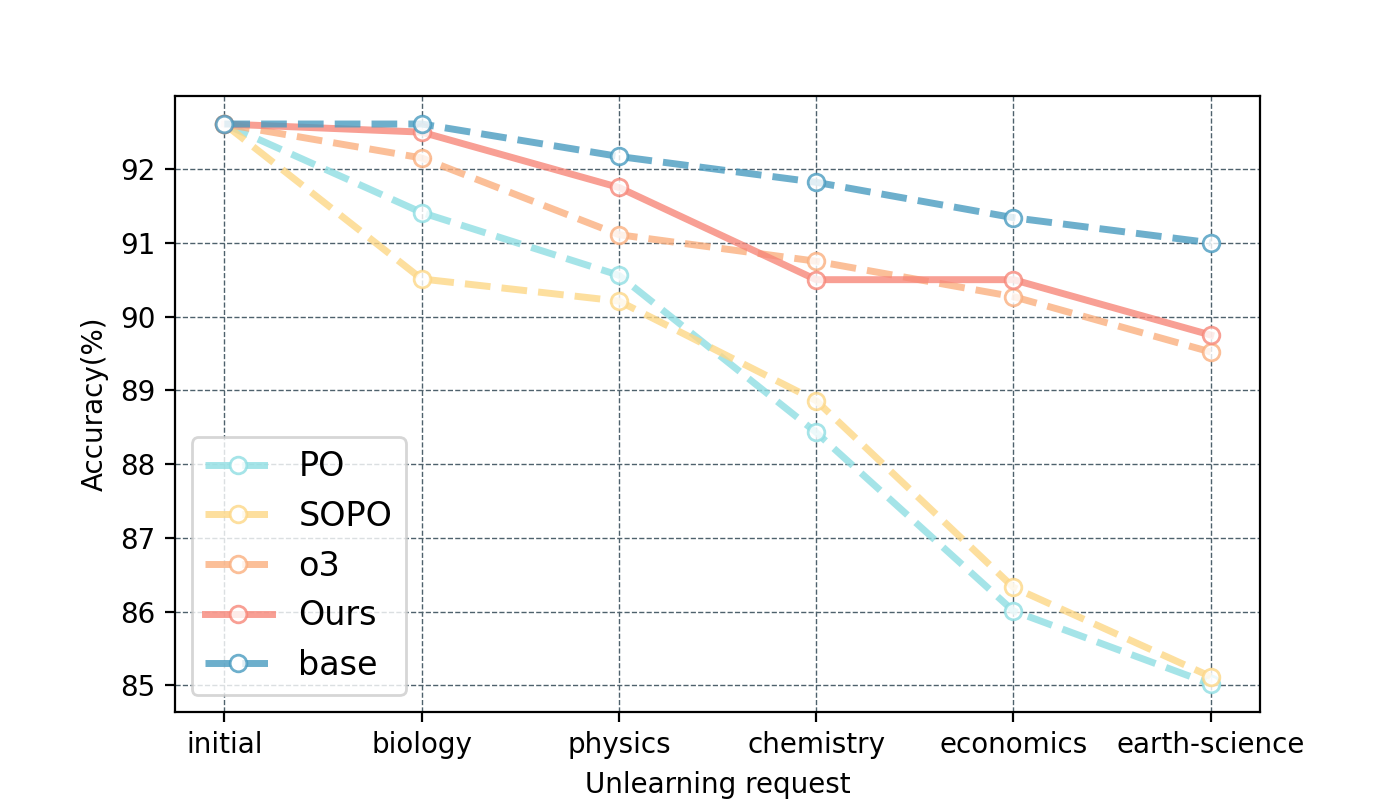}%
   \label{imagenetr-2}}

 \caption{The comparison results with other methods on the ScienceQA dataset.
 (a) The results of S.U..
 (b) The results of D.U..
 (c) The results of R.D..
}
\label{scqa-acc}
\end{figure*}



\section{Conclusion}




Existing LLM unlearning methods are vulnerable not only to cumulative catastrophic utility loss from continuous unlearning requests, but also to a fundamental limitation in practicality owing to their heavy reliance on the retained dataset.
To overcome this limitation, we introduce RCU, a retain-free approach that formulates LoRA updates as rotations within a specially constructed cognitive rotation space. 
This formulation makes the rotational angle updated by LoRA the sole variable, which is strongly correlated with the degree of unlearning.
Consequently, the unlearning process can be directly quantified by corresponding changes in the rotational angle.
Furthermore, by analyzing the update dynamics during unlearning, we propose a rotational salience weight to achieve precise and continuous control over the unlearning process.
Our method is supported by theoretical guarantees, and we rigorously establish its efficacy through a mathematical analysis.
Extensive experiments demonstrate that RCU achieves superior unlearning effectiveness while maintaining model utility on multiple datasets.


\begin{thebibliography}{47}
\providecommand{\natexlab}[1]{#1}
\providecommand{\url}[1]{\texttt{#1}}
\expandafter\ifx\csname urlstyle\endcsname\relax
  \providecommand{\doi}[1]{doi: #1}\else
  \providecommand{\doi}{doi: \begingroup \urlstyle{rm}\Url}\fi

\bibitem[Achiam et~al.(2023)Achiam, Adler, Agarwal, Ahmad, Akkaya, Aleman, Almeida, Altenschmidt, Altman, Anadkat, et~al.]{gpt}
Josh Achiam, Steven Adler, Sandhini Agarwal, Lama Ahmad, Ilge Akkaya, Florencia~Leoni Aleman, Diogo Almeida, Janko Altenschmidt, Sam Altman, Shyamal Anadkat, et~al.
\newblock Gpt-4 technical report.
\newblock \emph{arXiv preprint arXiv:2303.08774}, 2023.

\bibitem[Bourtoule et~al.(2021)Bourtoule, Chandrasekaran, Choquette-Choo, Jia, Travers, Zhang, Lie, and Papernot]{bourtoule}
Lucas Bourtoule, Varun Chandrasekaran, Christopher~A Choquette-Choo, Hengrui Jia, Adelin Travers, Baiwu Zhang, David Lie, and Nicolas Papernot.
\newblock Machine unlearning.
\newblock In \emph{2021 IEEE symposium on security and privacy (SP)}, pp.\  141--159. IEEE, 2021.

\bibitem[Bronec \& Helcl(2025)Bronec and Helcl]{atyaephyra}
Jan Bronec and Jind{\v{r}}ich Helcl.
\newblock Atyaephyra at semeval-2025 task 4: Low-rank negative preference optimization.
\newblock \emph{arXiv preprint arXiv:2503.13690}, 2025.

\bibitem[Cao et~al.(2024)Cao, Zhong, Zhou, Liu, Liu, and Han]{cao}
Chentao Cao, Zhun Zhong, Zhanke Zhou, Yang Liu, Tongliang Liu, and Bo~Han.
\newblock Envisioning outlier exposure by large language models for out-of-distribution detection.
\newblock \emph{arXiv preprint arXiv:2406.00806}, 2024.

\bibitem[Chen \& Yang(2023{\natexlab{a}})Chen and Yang]{EUL}
Jiaao Chen and Diyi Yang.
\newblock Unlearn what you want to forget: Efficient unlearning for llms.
\newblock \emph{arXiv preprint arXiv:2310.20150}, 2023{\natexlab{a}}.

\bibitem[Chen \& Yang(2023{\natexlab{b}})Chen and Yang]{canshuyouhua1}
Jiaao Chen and Diyi Yang.
\newblock Unlearn what you want to forget: Efficient unlearning for llms.
\newblock \emph{arXiv preprint arXiv:2310.20150}, 2023{\natexlab{b}}.

\bibitem[Chen et~al.(2025)Chen, Yao, Zhang, Shen, Liu, and Liu]{safety}
Yiwei Chen, Yuguang Yao, Yihua Zhang, Bingquan Shen, Gaowen Liu, and Sijia Liu.
\newblock Safety mirage: How spurious correlations undermine vlm safety fine-tuning.
\newblock \emph{arXiv preprint arXiv:2503.11832}, 2025.

\bibitem[Eldan \& Russinovich(2023)Eldan and Russinovich]{canshuyouhua2}
Ronen Eldan and Mark Russinovich.
\newblock Who’s harry potter? approximate unlearning for llms.
\newblock 2023.

\bibitem[Erfani et~al.(2016)Erfani, Rajasegarar, Karunasekera, and Leckie]{svmood}
Sarah~M Erfani, Sutharshan Rajasegarar, Shanika Karunasekera, and Christopher Leckie.
\newblock High-dimensional and large-scale anomaly detection using a linear one-class svm with deep learning.
\newblock \emph{Pattern Recognition}, 58:\penalty0 121--134, 2016.

\bibitem[Gallier(2001)]{basics-li}
Jean Gallier.
\newblock Basics of classical lie groups: The exponential map, lie groups, and lie algebras.
\newblock In \emph{Geometric Methods and Applications: For Computer Science and Engineering}, pp.\  367--414. Springer, 2001.

\bibitem[Gao et~al.(2025)Gao, Wang, Ding, Weng, Wang, and Zhu]{o3}
Chongyang Gao, Lixu Wang, Kaize Ding, Chenkai Weng, Xiao Wang, and Qi~Zhu.
\newblock On large language model continual unlearning.
\newblock In \emph{The Thirteenth International Conference on Learning Representations}, 2025.

\bibitem[Golatkar et~al.(2020)Golatkar, Achille, and Soatto]{gradasc}
Aditya Golatkar, Alessandro Achille, and Stefano Soatto.
\newblock Eternal sunshine of the spotless net: Selective forgetting in deep networks.
\newblock In \emph{Proceedings of the IEEE/CVF conference on computer vision and pattern recognition}, pp.\  9304--9312, 2020.

\bibitem[He et~al.(2025)He, Sarwar, Khalil, Yi, and Wang]{5deep}
Estrid He, Tabinda Sarwar, Ibrahim Khalil, Xun Yi, and Ke~Wang.
\newblock Deep contrastive unlearning for language models.
\newblock \emph{arXiv preprint arXiv:2503.14900}, 2025.

\bibitem[Zhao et~al.(2025)L Zhao, S Zalouk, CK Belardi, J Lovelace, JP Zhou, KQ Weinberger, Y Artzi, JJ Sun]{zhao1}
Linxi Zhao, Sofian Zalouk, Christian K. Belardi, Justin Lovelace, Jin Peng Zhou, Kilian Q. Weinberger, Yoav Artzi, Jennifer J. Sun.
\newblock Pre-training Large Memory Language Models with Internal and External Knowledge.
\newblock \emph{arXiv preprint arXiv:2505.15962}, 2025.

\bibitem[Daniele(2001)]{gangti}
Daniele Mortari.
\newblock On the rigid rotation concept in n-dimensional spaces.
\newblock In \emph{The Journal of the astronautical sciences}, pp.\ 401--420. Springer, 2001.



\bibitem[Hu et~al.(2022)Hu, Shen, Wallis, Allen-Zhu, Li, Wang, Wang, Chen, et~al.]{lora}
Edward~J Hu, Yelong Shen, Phillip Wallis, Zeyuan Allen-Zhu, Yuanzhi Li, Shean Wang, Lu~Wang, Weizhu Chen, et~al.
\newblock Lora: Low-rank adaptation of large language models.
\newblock \emph{ICLR}, 1\penalty0 (2):\penalty0 3, 2022.

\bibitem[Hu et~al.(2025)Hu, Huang, Yin, Ruan, Cheng, Dong, and Huang]{falcon}
Jinwei Hu, Zhenglin Huang, Xiangyu Yin, Wenjie Ruan, Guangliang Cheng, Yi~Dong, and Xiaowei Huang.
\newblock Falcon: Fine-grained activation manipulation by contrastive orthogonal unalignment for large language model.
\newblock \emph{arXiv preprint arXiv:2502.01472}, 2025.

\bibitem[Jia et~al.(2024)Jia, Zhang, Zhang, Liu, Runwal, Diffenderfer, Kailkhura, and Liu]{canshuyouhua3-soul}
Jinghan Jia, Yihua Zhang, Yimeng Zhang, Jiancheng Liu, Bharat Runwal, James Diffenderfer, Bhavya Kailkhura, and Sijia Liu.
\newblock Soul: Unlocking the power of second-order optimization for llm unlearning.
\newblock \emph{arXiv preprint arXiv:2404.18239}, 2024.

\bibitem[Jian et~al.(2022)Jian, Gao, and Vosoughi]{mlm}
Yiren Jian, Chongyang Gao, and Soroush Vosoughi.
\newblock Contrastive learning for prompt-based few-shot language learners.
\newblock \emph{arXiv preprint arXiv:2205.01308}, 2022.

\bibitem[Jiang et~al.(2025)Jiang, Lyu, Li, and Ma]{backdoor}
Peihai Jiang, Xixiang Lyu, Yige Li, and Jing Ma.
\newblock Backdoor token unlearning: Exposing and defending backdoors in pretrained language models.
\newblock In \emph{Proceedings of the AAAI Conference on Artificial Intelligence}, volume~39, pp.\  24285--24293, 2025.

\bibitem[Lang et~al.(2022)Lang, Zheng, Sun, Huang, Si, and Li]{lang2022estimating}
Hao Lang, Yinhe Zheng, Jian Sun, Fei Huang, Luo Si, and Yongbin Li.
\newblock Estimating soft labels for out-of-domain intent detection.
\newblock \emph{arXiv preprint arXiv:2211.05561}, 2022.

\bibitem[Laxhammar et~al.(2009)Laxhammar, Falkman, and Sviestins]{gaosimix}
Rikard Laxhammar, Goran Falkman, and Egils Sviestins.
\newblock Anomaly detection in sea traffic-a comparison of the gaussian mixture model and the kernel density estimator.
\newblock In \emph{2009 12th international conference on information fusion}, pp.\  756--763. IEEE, 2009.

\bibitem[Liu et~al.(2024)Liu, Feng, Xue, Wang, Wu, Lu, Zhao, Deng, Zhang, Ruan, et~al.]{deepseek}
Aixin Liu, Bei Feng, Bing Xue, Bingxuan Wang, Bochao Wu, Chengda Lu, Chenggang Zhao, Chengqi Deng, Chenyu Zhang, Chong Ruan, et~al.
\newblock Deepseek-v3 technical report.
\newblock \emph{arXiv preprint arXiv:2412.19437}, 2024.

\bibitem[Liu et~al.(2025)Liu, Yao, Jia, Casper, Baracaldo, Hase, Yao, Liu, Xu, Li, et~al.]{liu2025rethinking}
Sijia Liu, Yuanshun Yao, Jinghan Jia, Stephen Casper, Nathalie Baracaldo, Peter Hase, Yuguang Yao, Chris~Yuhao Liu, Xiaojun Xu, Hang Li, et~al.
\newblock Rethinking machine unlearning for large language models.
\newblock \emph{Nature Machine Intelligence}, pp.\  1--14, 2025.

\bibitem[Liu et~al.(2019)Liu, Ott, Goyal, Du, Joshi, Chen, Levy, Lewis, Zettlemoyer, and Stoyanov]{roberta}
Yinhan Liu, Myle Ott, Naman Goyal, Jingfei Du, Mandar Joshi, Danqi Chen, Omer Levy, Mike Lewis, Luke Zettlemoyer, and Veselin Stoyanov.
\newblock Roberta: A robustly optimized bert pretraining approach.
\newblock \emph{arXiv preprint arXiv:1907.11692}, 2019.

\bibitem[Lu et~al.(2022)Lu, Mishra, Xia, Qiu, Chang, Zhu, Tafjord, Clark, and Kalyan]{scienceqa}
Pan Lu, Swaroop Mishra, Tanglin Xia, Liang Qiu, Kai-Wei Chang, Song-Chun Zhu, Oyvind Tafjord, Peter Clark, and Ashwin Kalyan.
\newblock Learn to explain: Multimodal reasoning via thought chains for science question answering.
\newblock \emph{Advances in Neural Information Processing Systems}, 35:\penalty0 2507--2521, 2022.

\bibitem[Maini et~al.(2024)Maini, Feng, Schwarzschild, Lipton, and Kolter]{tofu}
Pratyush Maini, Zhili Feng, Avi Schwarzschild, Zachary~C Lipton, and J~Zico Kolter.
\newblock Tofu: A task of fictitious unlearning for llms.
\newblock \emph{arXiv preprint arXiv:2401.06121}, 2024.

\bibitem[Mihaylov et~al.(2018)Mihaylov, Clark, Khot, and Sabharwal]{suijisenlin}
Todor Mihaylov, Peter Clark, Tushar Khot, and Ashish Sabharwal.
\newblock Can a suit of armor conduct electricity? a new dataset for open book question answering.
\newblock \emph{arXiv preprint arXiv:1809.02789}, 2018.

\bibitem[Muhamed et~al.(2025)Muhamed, Bonato, Diab, and Smith]{saes}
Aashiq Muhamed, Jacopo Bonato, Mona~T Diab, and Virginia Smith.
\newblock Saes can improve unlearning: Dynamic sparse autoencoder guardrails for precision unlearning in llms.
\newblock In \emph{ICML 2025 Workshop on Reliable and Responsible Foundation Models}, 2025.

\bibitem[Ortiz-Jimenez et~al.(2023)Ortiz-Jimenez, Favero, and Frossard]{pan}
Guillermo Ortiz-Jimenez, Alessandro Favero, and Pascal Frossard.
\newblock Task arithmetic in the tangent space: Improved editing of pre-trained models.
\newblock \emph{Advances in Neural Information Processing Systems}, 36:\penalty0 66727--66754, 2023.

\bibitem[Ouyang et~al.(2023)Ouyang, Cao, Gao, Wu, Zhang, and Dai]{ouyang}
Yawen Ouyang, Yongchang Cao, Yuan Gao, Zhen Wu, Jianbing Zhang, and Xinyu Dai.
\newblock On prefix-tuning for lightweight out-of-distribution detection.
\newblock In \emph{Proceedings of the 61st Annual Meeting of the Association for Computational Linguistics (Volume 1: Long Papers)}, pp.\  1533--1545, 2023.

\bibitem[Pan et~al.(2020)Pan, Zhang, Ji, and Yang]{shangxiawenxiezai2}
Xudong Pan, Mi~Zhang, Shouling Ji, and Min Yang.
\newblock Privacy risks of general-purpose language models.
\newblock In \emph{2020 IEEE Symposium on Security and Privacy (SP)}, pp.\  1314--1331. IEEE, 2020.

\bibitem[Premptis et~al.(2025)Premptis, Lymperaiou, Filandrianos, Mastromichalakis, Voulodimos, and Stamou]{ails}
Iraklis Premptis, Maria Lymperaiou, Giorgos Filandrianos, Orfeas~Menis Mastromichalakis, Athanasios Voulodimos, and Giorgos Stamou.
\newblock Ails-ntua at semeval-2025 task 4: Parameter-efficient unlearning for large language models using data chunking.
\newblock \emph{arXiv preprint arXiv:2503.02443}, 2025.

\bibitem[Talmor et~al.(2018)Talmor, Herzig, Lourie, and Berant]{commonsenseqa}
Alon Talmor, Jonathan Herzig, Nicholas Lourie, and Jonathan Berant.
\newblock Commonsenseqa: A question answering challenge targeting commonsense knowledge.
\newblock \emph{arXiv preprint arXiv:1811.00937}, 2018.

\bibitem[Taori et~al.(2023)Taori, Gulrajani, Zhang, Dubois, Li, Guestrin, Liang, and Hashimoto]{openbookqa}
Rohan Taori, Ishaan Gulrajani, Tianyi Zhang, Yann Dubois, Xuechen Li, Carlos Guestrin, Percy Liang, and Tatsunori~B Hashimoto.
\newblock Alpaca: A strong, replicable instruction-following model.
\newblock \emph{Stanford Center for Research on Foundation Models. https://crfm. stanford. edu/2023/03/13/alpaca. html}, 3\penalty0 (6):\penalty0 7, 2023.

\bibitem[Thaker et~al.(2024)Thaker, Maurya, Hu, Wu, and Smith]{shangxiawenxiezai1}
Pratiksha Thaker, Yash Maurya, Shengyuan Hu, Zhiwei~Steven Wu, and Virginia Smith.
\newblock Guardrail baselines for unlearning in llms.
\newblock \emph{arXiv preprint arXiv:2403.03329}, 2024.

\bibitem[Touvron et~al.(2023)Touvron, Martin, Stone, Albert, Almahairi, Babaei, Bashlykov, Batra, Bhargava, Bhosale, et~al.]{llama7B-2}
Hugo Touvron, Louis Martin, Kevin Stone, Peter Albert, Amjad Almahairi, Yasmine Babaei, Nikolay Bashlykov, Soumya Batra, Prajjwal Bhargava, Shruti Bhosale, et~al.
\newblock Llama 2: Open foundation and fine-tuned chat models.
\newblock \emph{arXiv preprint arXiv:2307.09288}, 2023.

\bibitem[Wang et~al.(2024)Wang, Ma, Feng, Zhang, Yang, Zhang, Chen, Tang, Chen, Lin, et~al.]{wang-survey}
Lei Wang, Chen Ma, Xueyang Feng, Zeyu Zhang, Hao Yang, Jingsen Zhang, Zhiyuan Chen, Jiakai Tang, Xu~Chen, Yankai Lin, et~al.
\newblock A survey on large language model based autonomous agents.
\newblock \emph{Frontiers of Computer Science}, 18\penalty0 (6):\penalty0 186345, 2024.

\bibitem[Wang et~al.(2025{\natexlab{a}})Wang, Zhou, Zhou, Shin, Han, and Weinberger]{rethinking}
Qizhou Wang, Jin~Peng Zhou, Zhanke Zhou, Saebyeol Shin, Bo~Han, and Kilian~Q Weinberger.
\newblock Rethinking llm unlearning objectives: A gradient perspective and go beyond.
\newblock \emph{arXiv preprint arXiv:2502.19301}, 2025{\natexlab{a}}.

\bibitem[Wang et~al.(2025{\natexlab{b}})Wang, Zhang, Ye, Ren, Chen, and Ren]{uipe}
Wenyu Wang, Mengqi Zhang, Xiaotian Ye, Zhaochun Ren, Zhumin Chen, and Pengjie Ren.
\newblock Uipe: Enhancing llm unlearning by removing knowledge related to forgetting targets.
\newblock \emph{arXiv preprint arXiv:2503.04693}, 2025{\natexlab{b}}.

\bibitem[Xu et~al.(2021)Xu, Ren, Zhang, Feng, and Xiong]{xu-ood}
Keyang Xu, Tongzheng Ren, Shikun Zhang, Yihao Feng, and Caiming Xiong.
\newblock Unsupervised out-of-domain detection via pre-trained transformers.
\newblock \emph{arXiv preprint arXiv:2106.00948}, 2021.

\bibitem[Yang et~al.(2024)Yang, Zhou, Li, and Liu]{deepood}
Jingkang Yang, Kaiyang Zhou, Yixuan Li, and Ziwei Liu.
\newblock Generalized out-of-distribution detection: A survey.
\newblock \emph{International Journal of Computer Vision}, 132\penalty0 (12):\penalty0 5635--5662, 2024.

\bibitem[Yang et~al.(2025)Yang, Wang, Huang, Liu, Zhang, and Han]{exploring}
Puning Yang, Qizhou Wang, Zhuo Huang, Tongliang Liu, Chengqi Zhang, and Bo~Han.
\newblock Exploring criteria of loss reweighting to enhance llm unlearning.
\newblock \emph{arXiv preprint arXiv:2505.11953}, 2025.

\bibitem[Yao et~al.(2024)Yao, Xu, and Liu]{graddif}
Yuanshun Yao, Xiaojun Xu, and Yang Liu.
\newblock Large language model unlearning.
\newblock \emph{Advances in Neural Information Processing Systems}, 37:\penalty0 105425--105475, 2024.

\bibitem[Yi et~al.(2025)Yi, Huang, Zhang, Li, Nie, Liu, and Shen]{ctrap}
Biao Yi, Tiansheng Huang, Baolei Zhang, Tong Li, Lihai Nie, Zheli Liu, and Li~Shen.
\newblock Ctrap: Embedding collapse trap to safeguard large language models from harmful fine-tuning.
\newblock \emph{arXiv preprint arXiv:2505.16559}, 2025.

\bibitem[Yu et~al.(2025)Yu, Lin, Zhang, Li, Fang, Zhang, Wang, and Wang]{unierase}
Miao Yu, Liang Lin, Guibin Zhang, Xinfeng Li, Junfeng Fang, Ningyu Zhang, Kun Wang, and Yang Wang.
\newblock Unierase: Unlearning token as a universal erasure primitive for language models.
\newblock \emph{arXiv preprint arXiv:2505.15674}, 2025.

\bibitem[Zhang et~al.(2024)Zhang, Lin, Bai, and Mei]{NPO}
Ruiqi Zhang, Licong Lin, Yu~Bai, and Song Mei.
\newblock Negative preference optimization: From catastrophic collapse to effective unlearning.
\newblock \emph{arXiv preprint arXiv:2404.05868}, 2024.

\bibitem[Zhou et~al.(2023)Zhou, Yang, Wang, and Qiu]{zhou}
Yunhua Zhou, Jianqiang Yang, Pengyu Wang, and Xipeng Qiu.
\newblock Two birds one stone: Dynamic ensemble for ood intent classification.
\newblock In \emph{Proceedings of the 61st Annual Meeting of the Association for Computational Linguistics (Volume 1: Long Papers)}, pp.\  10659--10673, 2023.

\bibitem[Zong et~al.(2018)Zong, Song, Min, Cheng, Lumezanu, Cho, and Chen]{dagmm}
Bo~Zong, Qi~Song, Martin~Renqiang Min, Wei Cheng, Cristian Lumezanu, Daeki Cho, and Haifeng Chen.
\newblock Deep autoencoding gaussian mixture model for unsupervised anomaly detection.
\newblock In \emph{International conference on learning representations}, 2018.

\end{thebibliography}
\bibliographystyle{iclr2026_conference}

\appendix
\section{Appendix}

This Appendix includes additional details for the our paper including the following aspects:

• A.1: The proofs of the theorem 1 and theorem 2.

• A.2: Detailed experimental results on the ScienceQA dataset.

• A.3: Large Language Model usage declaration.

• A.4: The experimental results on the TOFU dataset showing the relationship between $\beta$ and the unlearning process.

• A.5: The detailed calculation steps of the OOD Score.

• A.6: An in-depth study on why our method is effective.

• A.7: Experimental results regarding the hyperparameter $\lambda$

• A.8: Computation overhead analysis.

• A.9: The comparison results of the Truth Ratio on the TOFU dataset.
\subsection{Proofs of the Theorems.}

\textbf{Theorem 1:} \textbf{\textit{For $R\in \mathbb{R} ^{n\times n} $, when $R=exp\left ( C \right ) $, the rotation angle $\theta$ of $R$ is directly proportional to $C$.}}

\textbf{Proof 1:} 
It is known that $R\in \mathit{SO}\left ( n \right ) $ is an n-dimensional rotation matrix ($n> 3$) and $R=exp\left ( C \right ) $ , where $C$ is an antisymmetric matrix.
If the rotation angle of $R=exp\left ( C \right ) $ is $\theta$ , then the rotation Angle of $exp\left ( kC \right ) $ is $k\theta$.

Due to $R\in \mathit{SO}\left ( n \right ) $, there exist orthogonal matrices $Q$ such that:
\begin{align}
R=Q\cdot diag\left ( 1,...,1.R\left ( \theta _{1}  \right ),...,R\left ( \theta _{m}  \right )  \right ) \cdot Q^{T} ,
  \label{LORAGENGXIN}
\end{align}
here, $R=\begin{bmatrix}
   \cos \theta _{j} &  -\sin \theta _{j} \\
  \sin \theta _{j}  & \cos \theta _{j}
\end{bmatrix}$ is Two-dimensional rotation matrix (the rotation angle is $\theta _{j} $ ), the rest of the eigenvalues are 1. 
The $m=\left [ n/2 \right ] $.

The antisymmetric matrix $C$ can be similarly block-diagonalized as follows:
\begin{align}
C=Q\cdot diag\left ( 0,...,0,B\left ( \theta _{1}  \right ),..., B\left ( \theta _{m}  \right ) \right )\cdot Q^{T} ,
  \label{LORAGENGXIN}
\end{align}
where $B=\begin{bmatrix}
  0 &  -\theta _{j} \\
   \theta _{j}  & 0
\end{bmatrix}$ .
$exp\left ( B\left ( \theta _{j}  \right )  \right )  =R\left ( \theta _{j}  \right ) $.

If $C'=kC$ ,we can gather:
\begin{align}
  C'=Q\cdot diag\left ( 0,...,0,B\left ( \theta _{1}  \right ),..., B\left ( \theta _{m}  \right ) \right )\cdot Q^{T} ,
  \label{LORAGENGXIN}
\end{align}
since B is linear, $B\left ( k\theta _{j}  \right ) = k B\left ( \theta _{j}  \right )$.
Then:
\begin{equation}
   \begin{aligned}
    exp\left ( kC \right ) &= exp\left ( C' \right ) \\
    &= Q\cdot e^ {\left ( diag\left ( 0,...,0,B\left ( k\theta _{1}  \right ),..., B\left ( k\theta _{m}  \right ) \right ) \right )  }\cdot Q^{T} \\
    &=Q\cdot diag\left ( 1,...,1.B\left ( k\theta _{1}  \right ),...,B\left ( k\theta _{m}  \right )  \right ) \cdot Q^{T}.
    \end{aligned}
  \label{1-12}
\end{equation}

And the  $exp \left ( B\left ( k\theta _{j}  \right ) \right ) $ is as follow:
\begin{equation}
exp \left ( B\left ( k\theta _{j}  \right ) \right ) = \begin{bmatrix}
   \cos \left ( k\theta _{j}  \right ) & - \sin \left ( k\theta _{j}  \right )\\
   \sin \left ( k\theta _{j}  \right ) & \cos \left ( k\theta _{j}  \right )
   \end{bmatrix} =R\left ( k\theta _{j}  \right ) .
  \label{ZHENG3}
\end{equation}

Due to the \autoref{1-12} and \autoref{ZHENG3}, we can get as follow:
\begin{align}
exp \left ( kC \right ) = Q\cdot diag\left ( 0,...,0,R\left ( k\theta _{1}  \right ),...,R\left ( k\theta _{m}  \right ) \right )\cdot Q^{T},
  \label{ZHENG4}
\end{align}
here means the rotation angle of $exp \left ( kC \right ) $ are $k\theta _{1},...,k\theta _{m}$.

\textbf{Theorem 2:} 
when $R=exp\left ( A \right )$ and $R'=exp\left ( A' \right )$ , $A\bot A'$, then the rotation axes of $R$ and $R'$ are perpendicular to each other.

\textbf{Proof 2:}
Let $A$ and $A'$ be skew-symmetric matrices with their rotation faces $P$ and $P'$, respectively. 
Assume that $P$ and $P'$ are orthogonal to each other ($A$ and $A'$ are perpendicular).
Since $P\bot P'$, there are $P\subseteq ker\left ( A' \right )$ and $P'\subseteq ker\left ( A \right )$. 
(Here, $ker\left ( \cdot  \right ) $ refers to the null space.)
Thus, for any vector $v$, $A'Av=0$ and $AA'v=0$ , which is $AA' = A'A =0$.

Now, $R=exp\left ( A \right ) =I+A+\frac{A^2}{2!}+...$.
Similarly, $R'=exp\left ( A' \right )$.

The rotation face of $R$ is $P$ (the eigenspace corresponding to the nonzero eigenvalues of $A$), since $exp\left ( A \right )$ is the usual rotation on $P$ and identity elsewhere.
Similarly, the surface of rotation of $R'$ is $P'$.
Since $P$ and $P'$ are orthogonal, the spaces of rotation of $R$ and $R'$ are perpendicular to each other.

Thus, when the rotation faces of the skewsymmetric matrices $A$ and $A'$ generating rotations are perpendicular to each other, the rotation spaces of the corresponding rotation matrices $R'$ and $R$ are also perpendicular to each other.

\subsection{More Experimental Results}

The detailed results of our work on the ScienceQA dataset are shown in \autoref{SQA-21}.
Based on the results, we can observe that compared to $o^3$, our method is more stable and has a better ability to resist cumulative catastrophic utility loss.
Here, the $U^2R$ of our method is 65.79, and the $U^2R$ of the other method is 48.46.
\begin{table}[htbp]
 \caption{Performance Comparison between our method and other baselines when continually unlearning biology, physics, chemistry, economics and earth-science in Fictitious Knowledge Generation. 
 The unlearning effectiveness is measured by the generation accuracy of the unlearning train data and unlearning test data denoted as S.U. and D.U., CommonsenseQA (C.QA.), OpenbookQA (O.QA.) respectively. 
 Utility preservation is evaluated by the generation accuracy of Retained Distribution (R.D.).
  The * represents the results we achieved in our own experimental environment.}
\renewcommand\arraystretch{1}
  \centering
  \resizebox{1\columnwidth}{!}{
  \begin{tabular}{c|ccccc|ccccc|ccccc|}
    \toprule
      Method & \multicolumn{5}{c}{Unlearning Request 1}& \multicolumn{5}{c}{Unlearning Request 2}& \multicolumn{5}{c}{Unlearning Request 3} \\
          &  S.U.$\downarrow$ & D.U.$\downarrow$ & R.D.$\uparrow$ & C.QA.$\uparrow$ & O.QA.$\uparrow$ &  S.U.$\downarrow$ & D.U.$\downarrow$ & R.D.$\uparrow$ & C.QA.$\uparrow$ & O.QA.$\uparrow$  &  S.U.$\downarrow$ & D.U.$\downarrow$ & R.D.$\uparrow$ & C.QA.$\uparrow$ & O.QA.$\uparrow$  \\
    \midrule
    Base * &100&99.24&92.61&79.19&83.2&99.66&98.3&92.17&79.192&83.2&99.73&98.02&91.82&79.19&83.2\\
    \midrule
    $o^3$ *  &9.73&27.4&92.15&79.19&83.0&4.19&20.03&91.11&79.19&83.0&4.38&20.48&90.25&79.19&83.0 \\
    \rowcolor{gray!20}
    Ours  &\textbf{0.05}&\textbf{0.12}&\textbf{92.5}&\textbf{79.19}&\textbf{83.2}&\textbf{0.17}&\textbf{2.45}&\textbf{91.75}&\textbf{79.19}&\textbf{83.2}&\textbf{0.33}&\textbf{6.13}&\textbf{90.5}&\textbf{79.19}&\textbf{83.2} \\
        
    \bottomrule
  \end{tabular}
  }

\renewcommand\arraystretch{1}
  \centering
  \resizebox{0.7\columnwidth}{!}{
  \begin{tabular}{c|ccccc|ccccc|}
    \toprule
      Method & \multicolumn{5}{c}{Unlearning Request 4}& \multicolumn{5}{c}{Unlearning Request 5}\\
           &  S.U.$\downarrow$ & D.U.$\downarrow$ & R.D.$\uparrow$ & C.QA.$\uparrow$ & O.QA.$\uparrow$ &  S.U.$\downarrow$ & D.U.$\downarrow$ & R.D.$\uparrow$& C.QA.$\uparrow$ & O.QA.$\uparrow$  \\
    \midrule
    Base &99.75&98.23&91.34&79.19&83.2&99.77&98.25&91.0&79.19&83.2\\
    \midrule
    $o^3$  &8.07&23.98&90.17&79.19&83.0&11.8&28.0&89.52&79.19&83.0 \\
    
    \rowcolor{gray!20}
    Ours  &\textbf{0.49}&\textbf{9.56}&\textbf{90.5}&\textbf{79.19}&\textbf{83.2}&\textbf{0.98}&\textbf{10.29}&\textbf{89.753}&\textbf{79.19}&\textbf{83.2}\\
        
    \bottomrule
  \end{tabular}
  }
  \label{SQA-21}
\end{table}

\begin{figure*}[t]
\centering
\subfloat[]{\includegraphics[width=2.5in]{ 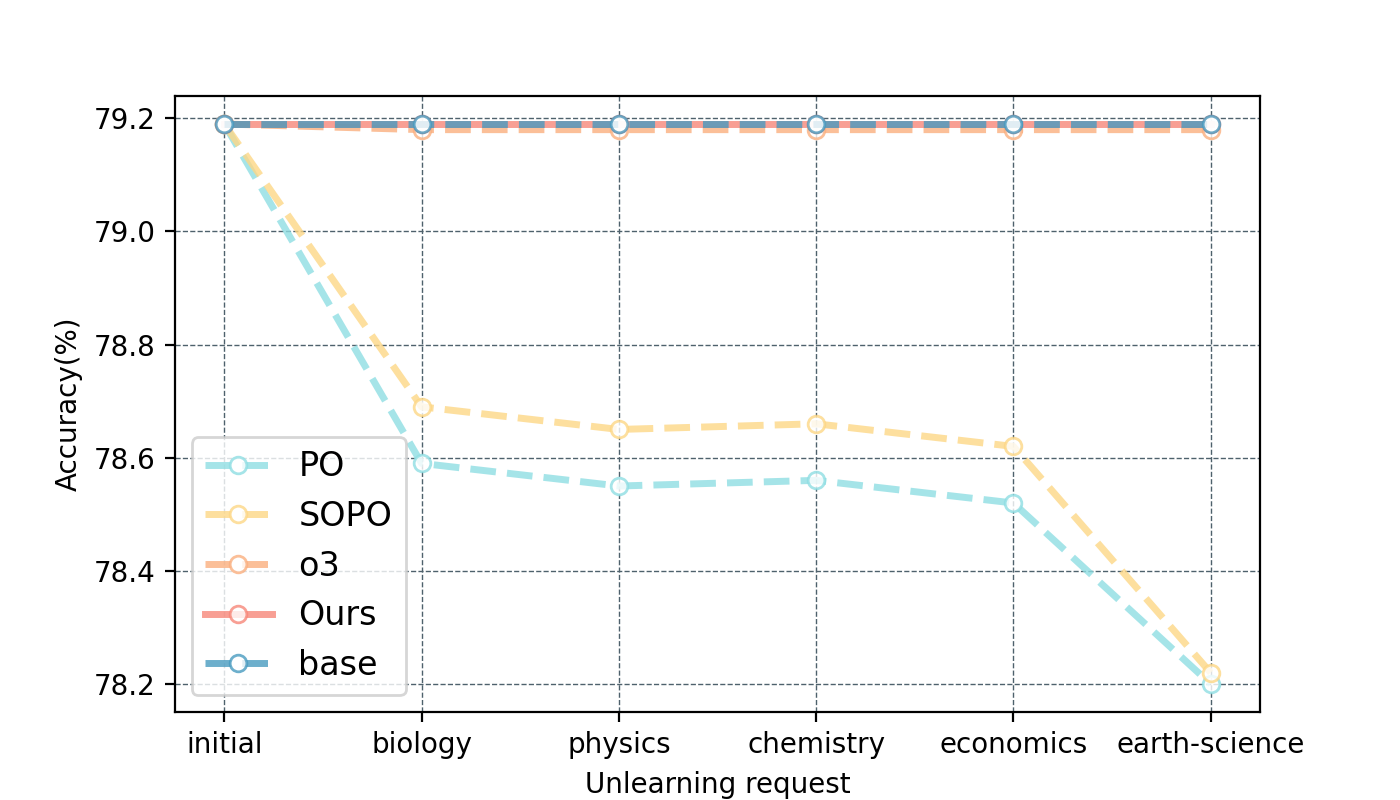}
   \label{imagenetr-2}}
\hfil
\subfloat[]{\includegraphics[width=2.5in]{ 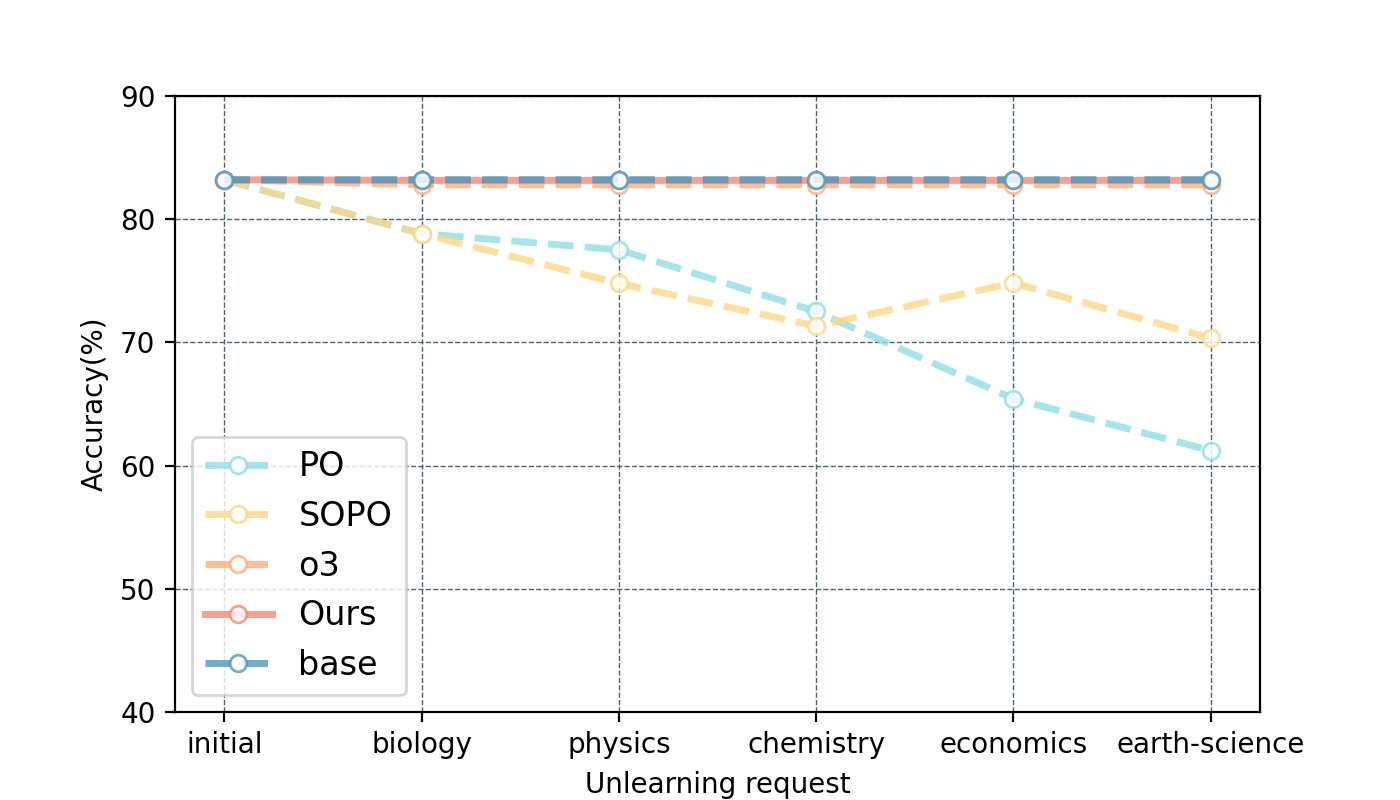}%
\label{imagenetr-2}}
 \caption{Experimental results on the ScienceQA dataset.
    (a) The results of C.QA..
    (b)The results of O.QA..
    }
\label{SC-QA}
\end{figure*}

The specific results are shown in \autoref{SC-QA}.

\subsection{Large Language Model Usage Declaration}
During this research process, the LLM provided significant assistance in organizing the logic of the paper and improving the language expression. 
Here, we express our gratitude for the role that the LLM played in enhancing the logicality and clarity of this research.

\subsection{Hyperparameter analysis on the TOFU dataset}
\begin{figure*}[h]
\centering
\subfloat[]{\includegraphics[width=2.5in]{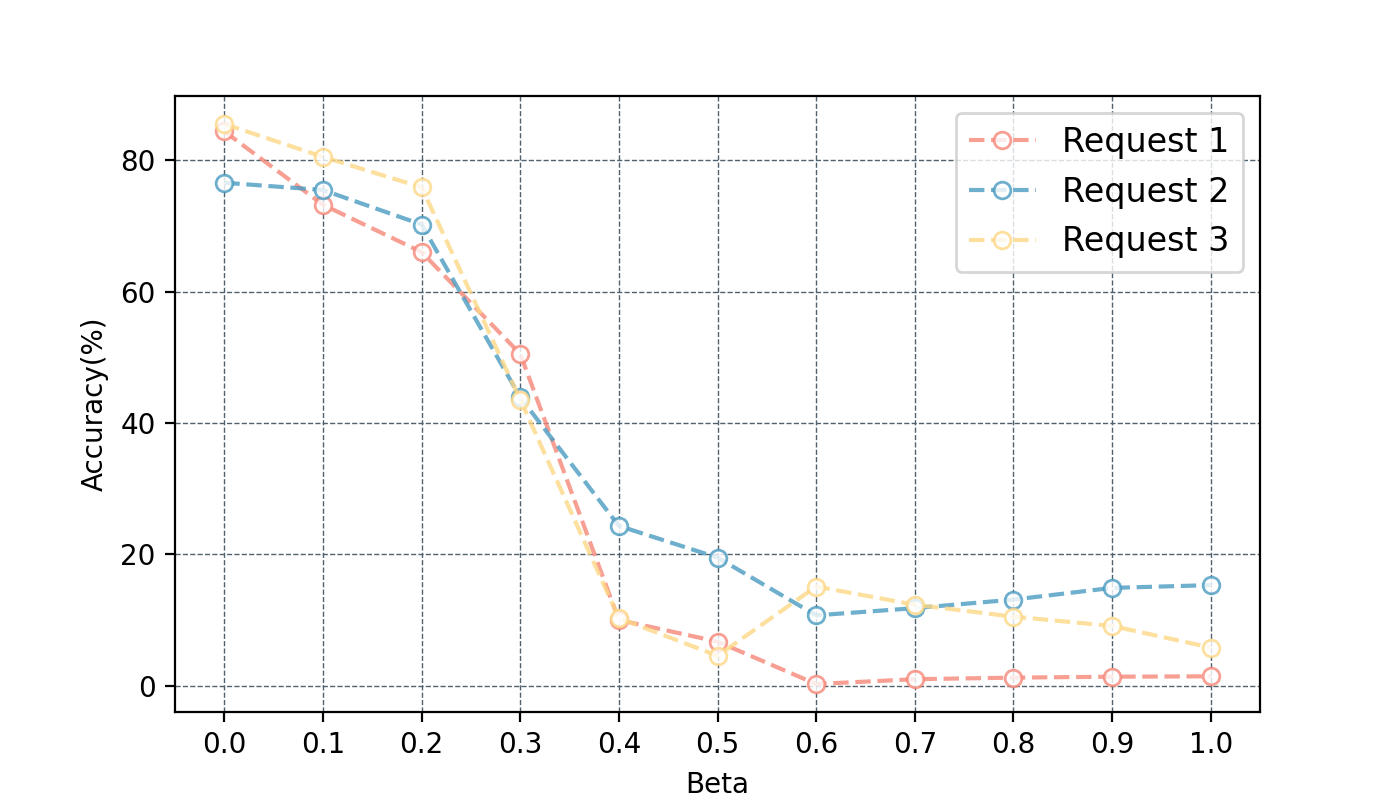}%
   \label{imagenetr-1}}
\hfil
\subfloat[]{\includegraphics[width=2.5in]{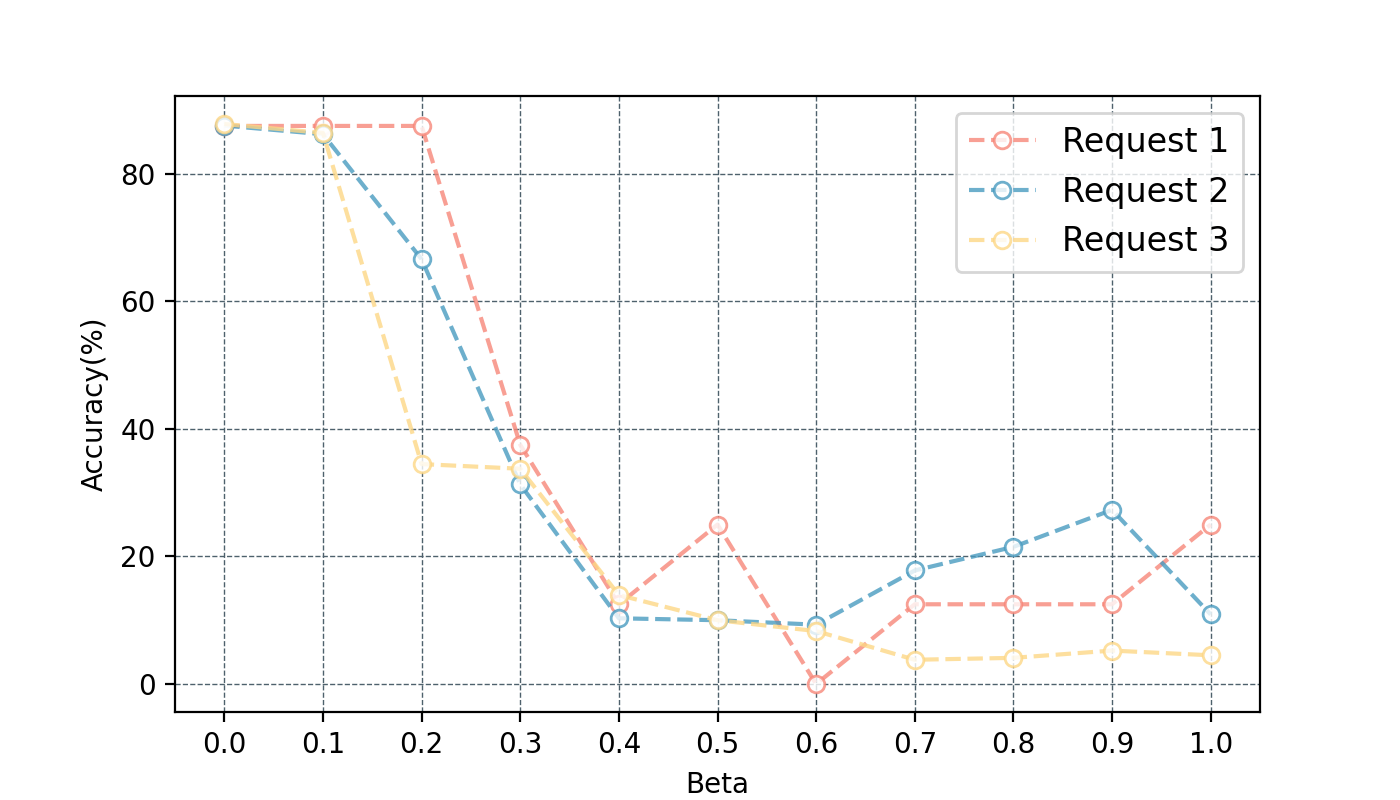}%
   \label{imagenetr-2}}
 \caption{The relationship between the $\beta$ process and the unlearning process.
 (a) The results of S.U. on the TOFU dataset.
    (b) The results of D.U. on the TOFU dataset.
    }
\label{result-tofu}
\end{figure*}

The results on TOFU dataset are shown in \autoref{result-tofu}. 
We found that the change of $\beta$ shifted slightly to the right for a short period, but the overall change was still concentrated in a certain area.
Therefore, when designing the distributional shift compensator for TOFU dataset, we also tried to map $\mathcal{M}\left ( \gamma^t  \right ) $ to the range of 0.2 to 0.6.

\subsection{Calculation of OOD Score}
We follow the method for obtaining the output of OOD detection as described in $o^3$. 
Our method for calculating the OOD score is as follows:
\begin{align}
s\left ( x \right ) _{l}=\left ( f_{w\left [ 1:l \right ] } -\mu _{l}  \right )^{T}    {\textstyle \sum_{l}^{-1}} \left ( f_{w\left [ 1:l \right ] } -\mu _{l}  \right ) +\gamma \cdot \left ( -\max_{i=1}^{\alpha N^{U} }\left \{ \frac{f_{w_{\left [ 1:l \right ]}}\left ( x \right )\cdot f_{w_{\left [ 1:l \right ]}}\left ( x_{i}^{U}  \right ) }{\left | f_{w_{\left [ 1:l \right ]}}\left ( x \right ) \right | \left | f_{w_{\left [ 1:l \right ]}}\left ( x_{i}^{U}  \right ) \right | }  \right \}   \right ) ,
  \label{LORAGENGXIN}
\end{align}
\begin{align}
\mu_{l}=\frac{1}{\alpha N^{\mathrm{U}}} \sum_{i=1}^{\alpha N^{\mathrm{U}}} f_{\omega_{[1: l]}}\left(\boldsymbol{x}_{i}^{\mathrm{U}}\right), \Sigma_{l}=\frac{1}{\alpha N^{\mathrm{U}}} \sum_{i=1}^{\alpha N^{\mathrm{U}}}\left(f_{\omega_{[1: l]}}\left(\boldsymbol{x}_{i}^{\mathrm{U}}\right)-\mu_{l}\right)\left(f_{\omega_{[1: l]}}\left(\boldsymbol{x}_{i}^{\mathrm{U}}\right)-\mu_{l}\right)^{\top}  ,
  \label{LORAGENGXIN}
\end{align}
here, $r = 1000$, $f_{w_{l}}$representing the parameter of layer $l$. 
$D_{used}^{U,t}$ refers to one of the two subsets randomly divided from the training dataset $D^{U,t}$, which contains $\alpha N^{U,t}$ samples.

When the $T$-th re-learning request is completed, each test input $x$ is input into the OOD detection to calculate the score vector, and the distance between $x$ and the hyper-spherical $\mathcal{H} ^{t} \left ( c^{t},r^{t} \right )$ boundary is obtained using one-class SVM. 
The final score is:
\begin{align}
d_{\mathcal{H^{t} } } \left ( x \right ) =\left | s\left ( x \right ) ^{t} -c^{t}  \right | -r^{t} ,
\label{LORAGENGXIN}
\end{align}
\begin{align}
\gamma ^{t}=\delta\left\{\zeta\left[1-\max \left(p, p^{\prime}\right)+\min \left(p, p^{\prime}\right)\right]\right\},  p=\mathcal{P}_{\text {mix }}^{t}\left(\mathrm{~d}_{\mathcal{H}^{t}}(\boldsymbol{x})\right), p^{\prime}=\mathcal{P}_{\text {mix }}^{t}\left(2 \mathrm{~d}_{\mathcal{H}^{t}}^{0}-\mathrm{d}_{\mathcal{H}^{t}}(\boldsymbol{x})\right),
  \label{LORAGENGXIN}
\end{align}
here, $c^{t}$ and $r^t$ represent the center vector and radius of the hypersphere. 
$\mathcal{P}_{\text {mix }}$ is the mixed gaussian distribution function.

\subsection{Research on Our Method}
\begin{wraptable}{r}{8.0cm}
	 \caption{
     On the ScienceQA dataset, the parameter $\Theta_{BA}$ maintained a stable performance at the $10^{-6}$ scale in multiple unlearning requests (block 1 in attention layers)
 }
\renewcommand\arraystretch{1}
  \centering
  \begin{tabular}{c|cc}
    \toprule
      Unlearning Request & Param $B$ & Param $A$  \\
    \midrule
    biology &$10^{-6}$&$10^{-5}$\\
    physics  &$10^{-5}$&$10^{-5}$ \\
    chemistry &$10^{-5}$&$10^{-5}$ \\
    economics &$10^{-6}$&$10^{-5}$\\
    earth-science &$10^{-6}$&$10^{-5}$\\
    \bottomrule
  \end{tabular}
  \label{parm}
\end{wraptable}
Starting from the intrinsic characteristics of the model parameters, we analyzed the proposed method. 
Specifically, the magnitude of the learnable parameter $\Theta_{BA}$ in each training round is shown in \autoref{parm}. 
The study found that when the parameters $\Theta_{BA}<<1$, the matrix $I + BA$ always has a corresponding cognitive rotation space.
In the ablation experiments (\autoref{AB}), when we remove the designed LoRA update paradigm, the unlearning effect measured by D.U. is relatively ideal when the unlearning request is 1 to 2.
However, when the unlearning request reached 3 or higher, due to the cumulative catastrophic utility loss, the unlearning performance reflected by D.U. significantly decreased.
The core of this method lies in introducing the Lsk loss, which can constrain the parameters $\Theta_{BA}$ to continuously maintain a small amplitude, thereby maintaining the effectiveness of the cognitive rotation space and ensuring the continuous efficacy of the method in handling the continuous unlearning.
The results shown in \autoref{parm} pertain to the queries layer of the attention layers (block 1).
The other results generally vary within the range of $10^{-6}$ to $10^{-5}$.

\begin{figure*}[h]
\centering
\subfloat[]{\includegraphics[width=2.5in]{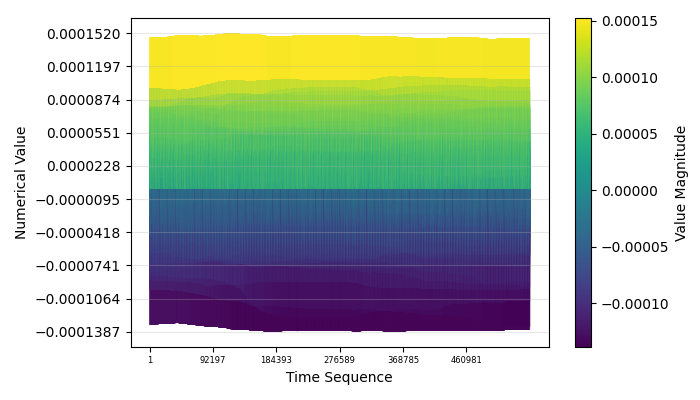}%
   \label{imagenetr-1}}
\hfil
\subfloat[]{\includegraphics[width=2.5in]{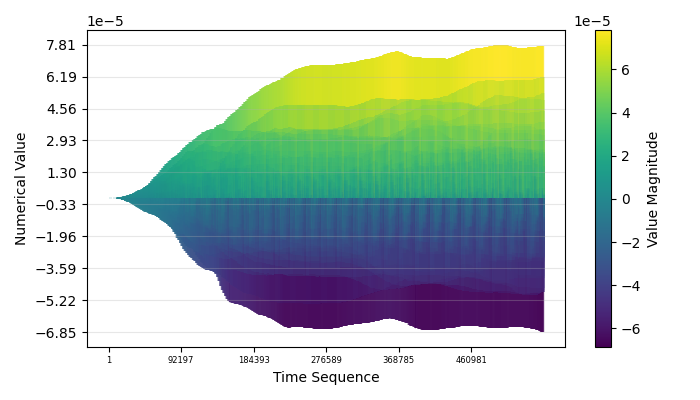}%
   \label{imagenetr-2}}
 \caption{The results of 5 continuous unlearning processes on the ScienceQA dataset.
 (a) The changes in the $\Theta_{A}$.
    (b) The changes in the $\Theta_{B}$.
    }
\label{LORAba}
\end{figure*}

Furthermore, \autoref{LORAba} (a)(b) present the evolution of parameters $\Theta_{A}$ and $\Theta_{B}$ throughout the experiment. 
The results indicate that the magnitudes of both $\Theta_{A}$ and $\Theta_{B}$ remain significantly smaller than those of parameter I during the entire training process.

\subsection{Hyperparameter Analysis}

\begin{figure*}[h]
\centering
\subfloat[]{\includegraphics[width=1.8in]{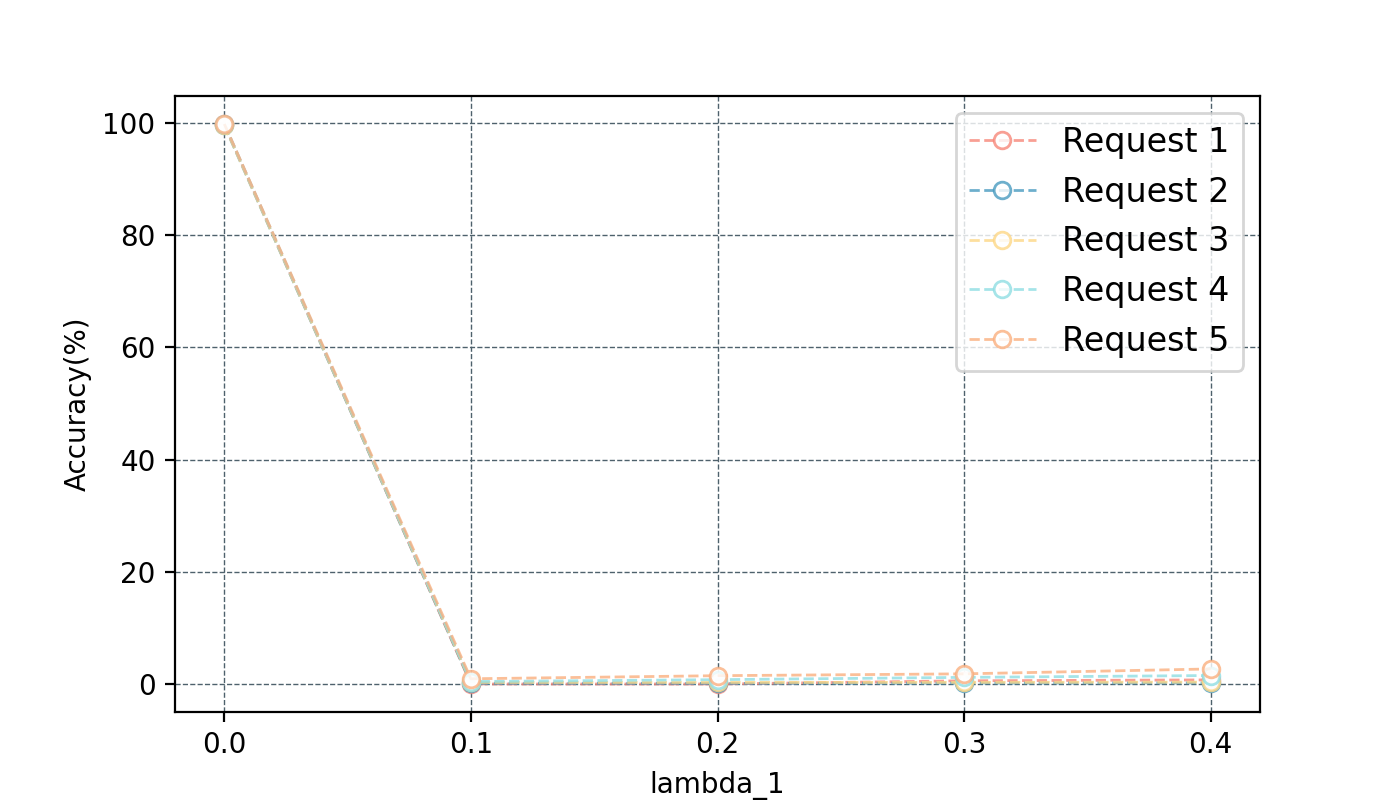}%
   \label{imagenetr-1}}
\hfil
\subfloat[]{\includegraphics[width=1.8in]{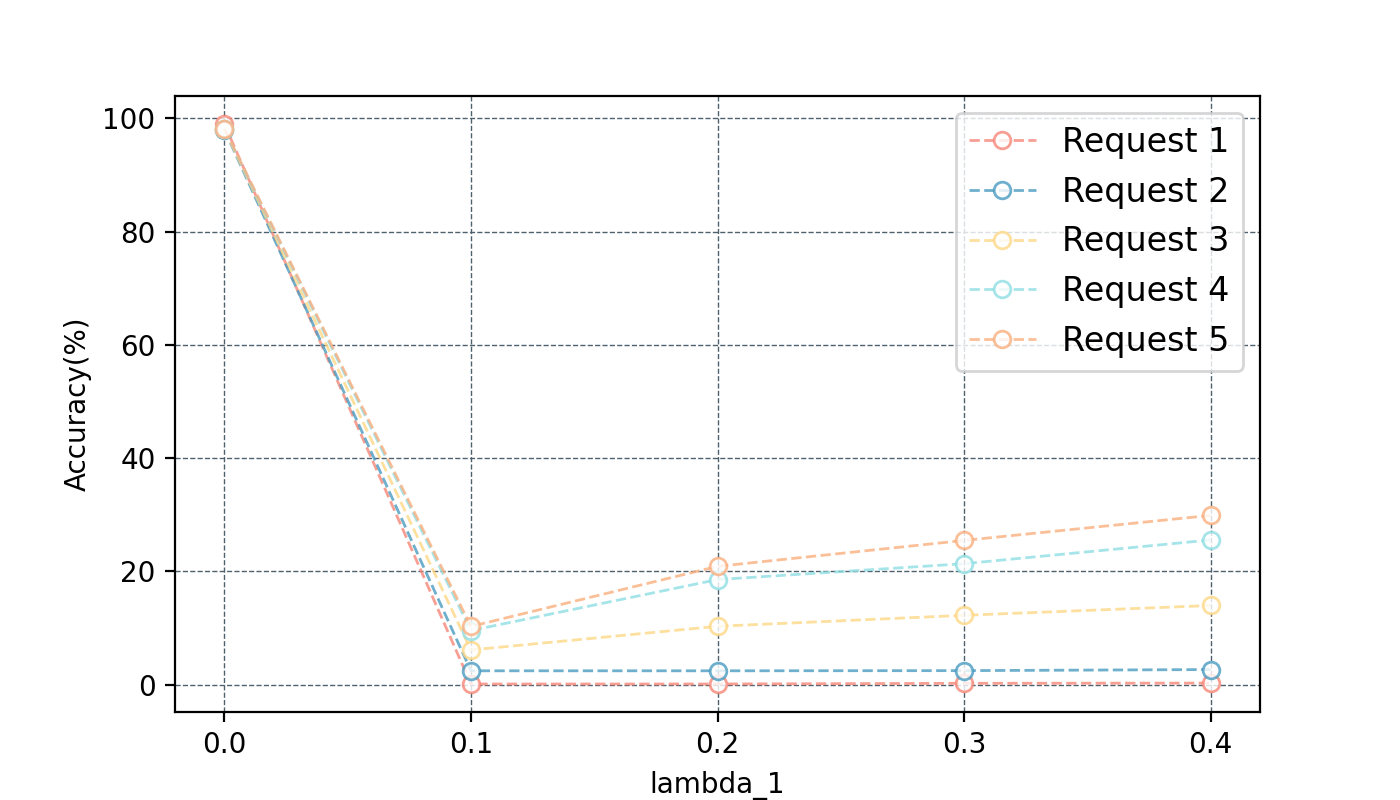}%
   \label{imagenetr-2}}
   \hfil
\subfloat[]{\includegraphics[width=1.8in]{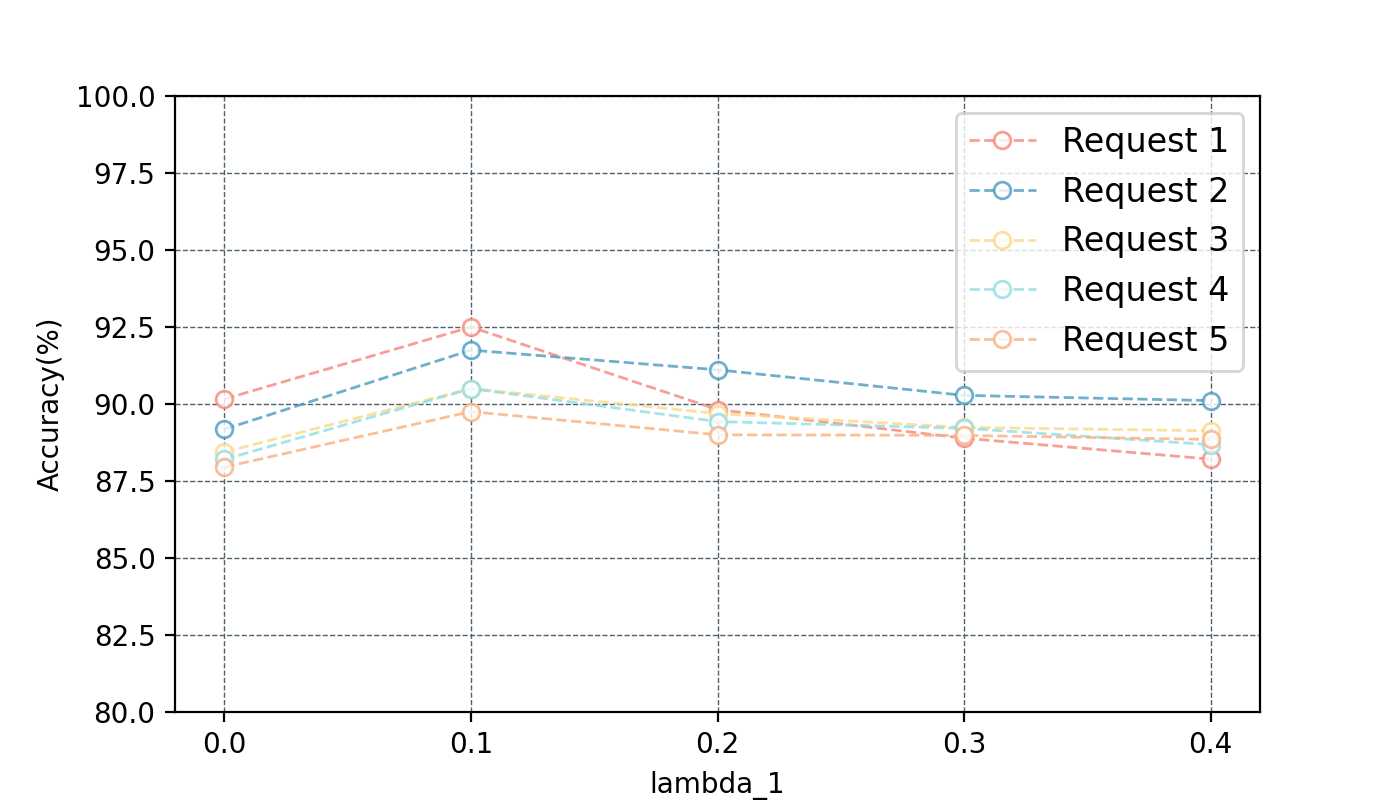}
   \label{imagenetr-2}}
  \hfil
\subfloat[]{\includegraphics[width=1.8in]{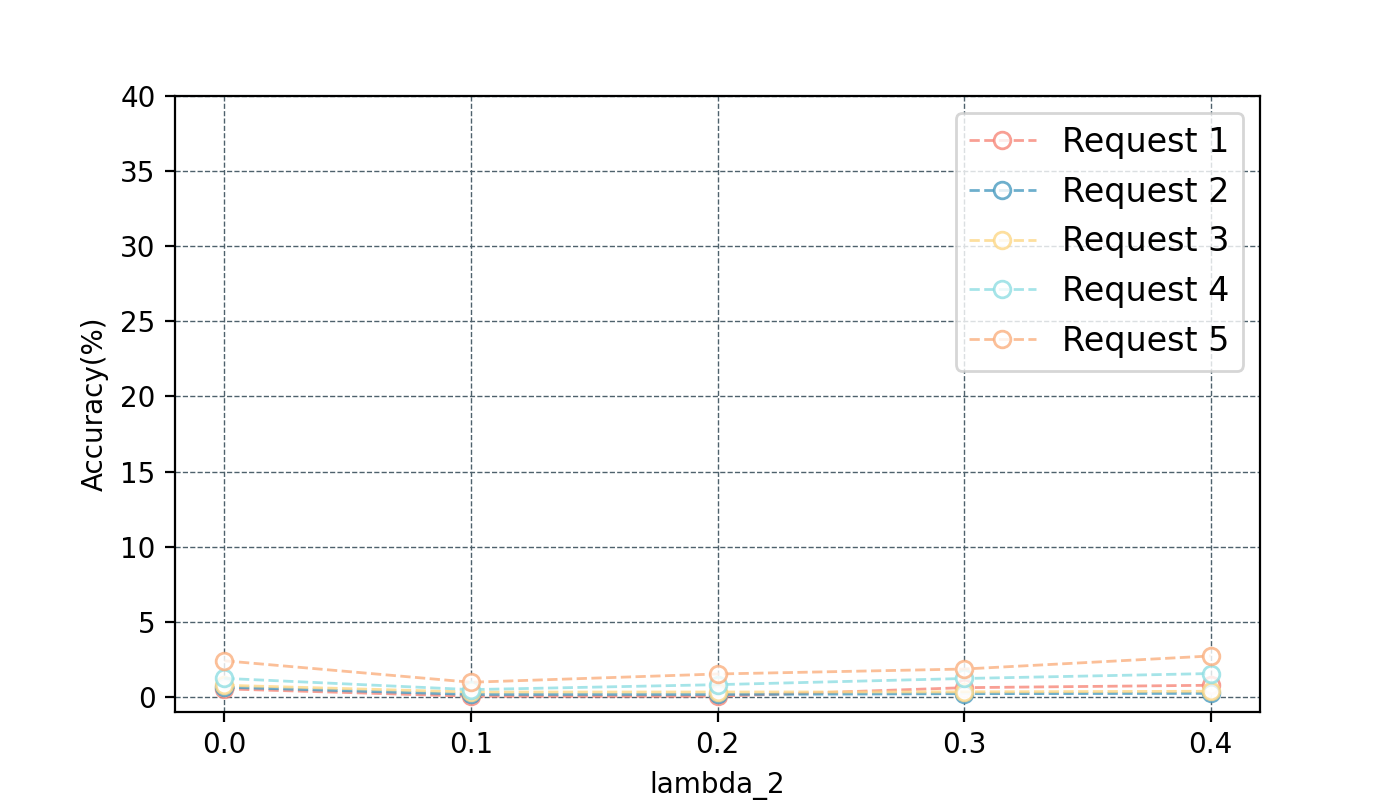}%
\label{imagenetr-2}}
  \hfil
\subfloat[]{\includegraphics[width=1.8in]{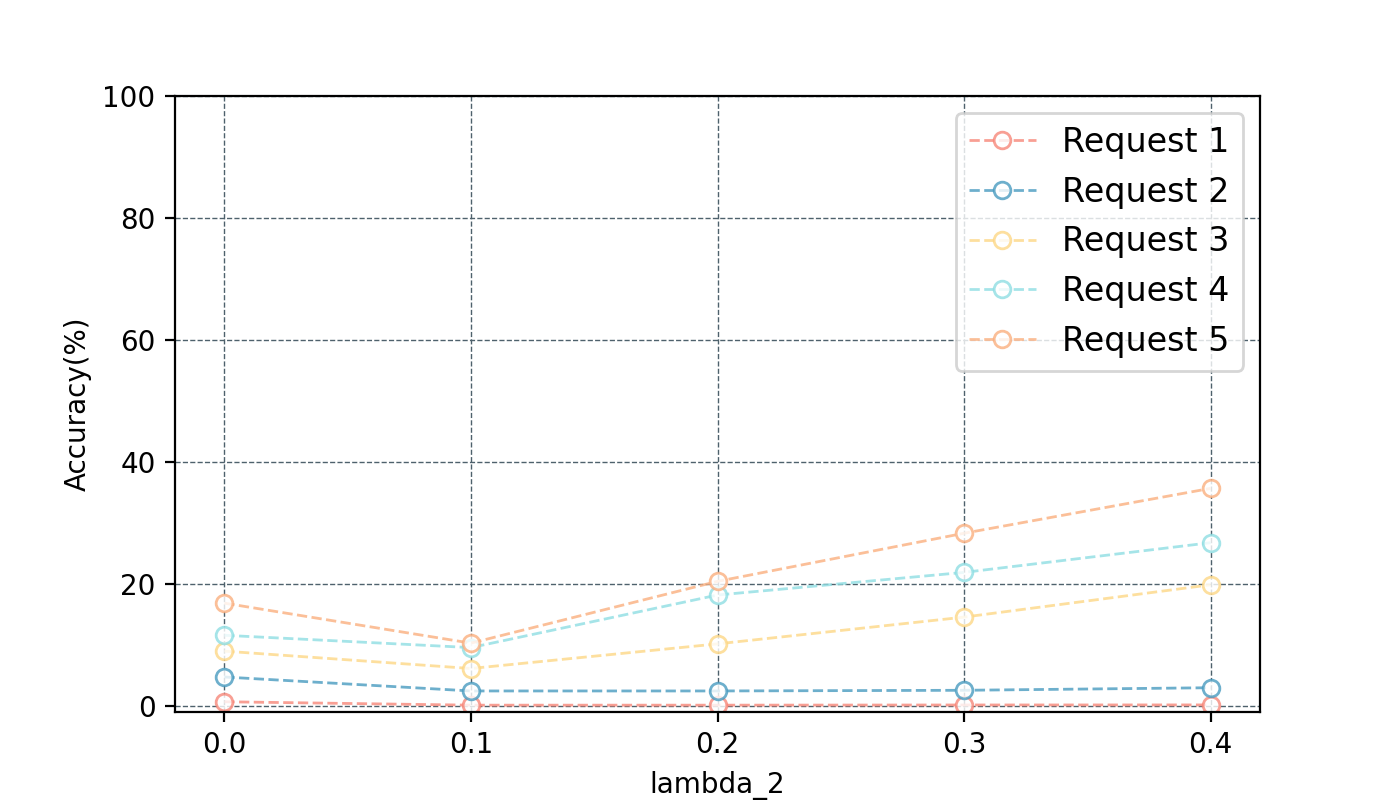}%
   \label{imagenetr-2}}
   \hfil
\subfloat[]{\includegraphics[width=1.8in]{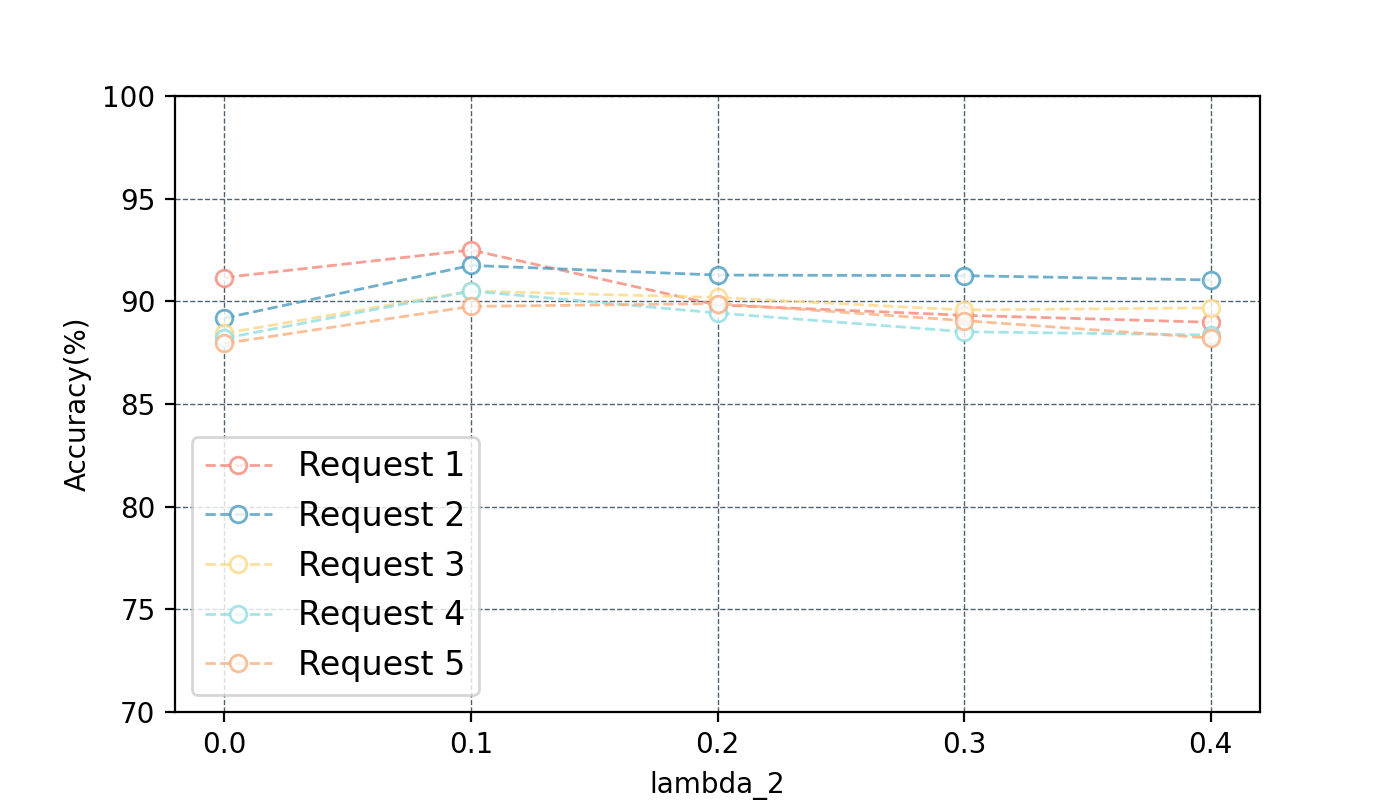}%
   \label{imagenetr-2}}
    \hfil
\subfloat[]{\includegraphics[width=1.8in]{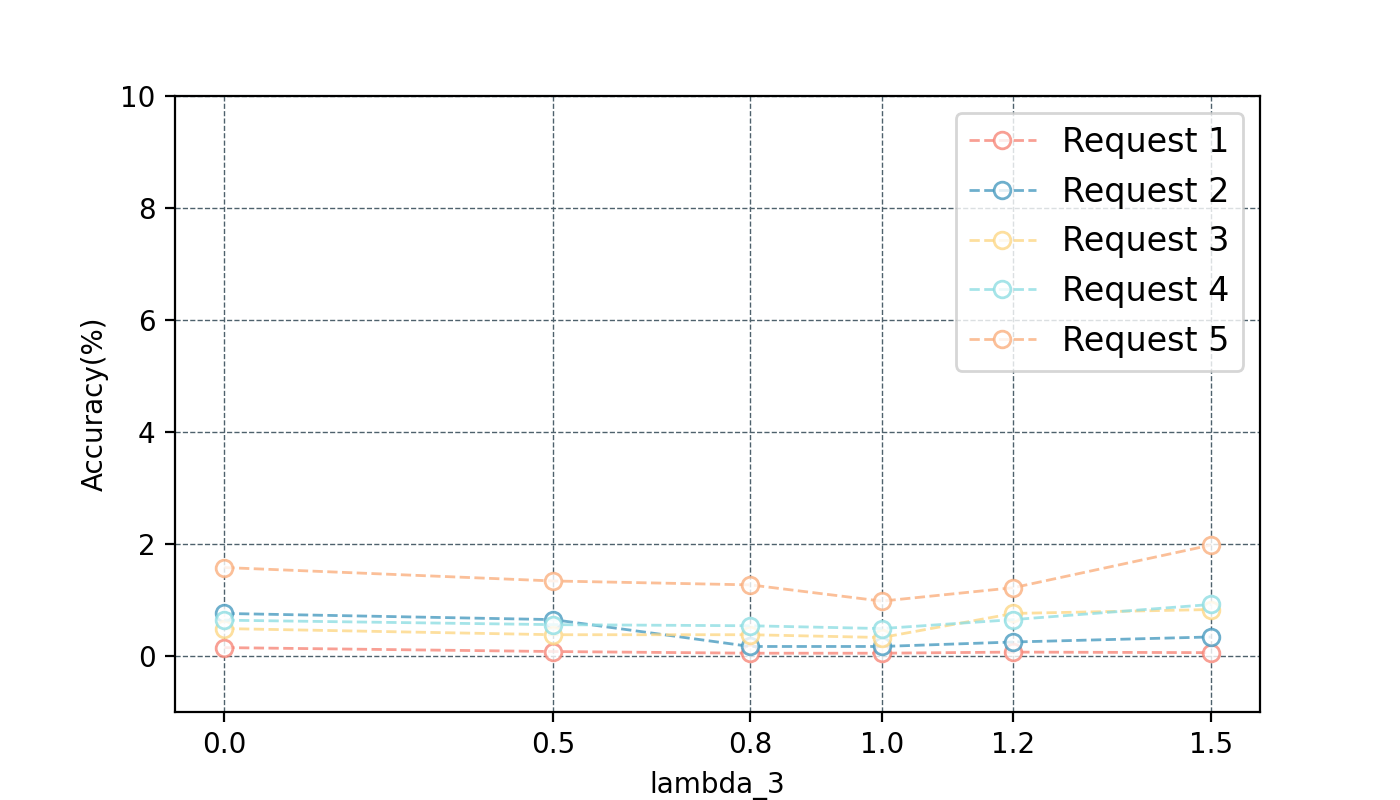}%
\label{imagenetr-2}}
  \hfil
\subfloat[]{\includegraphics[width=1.8in]{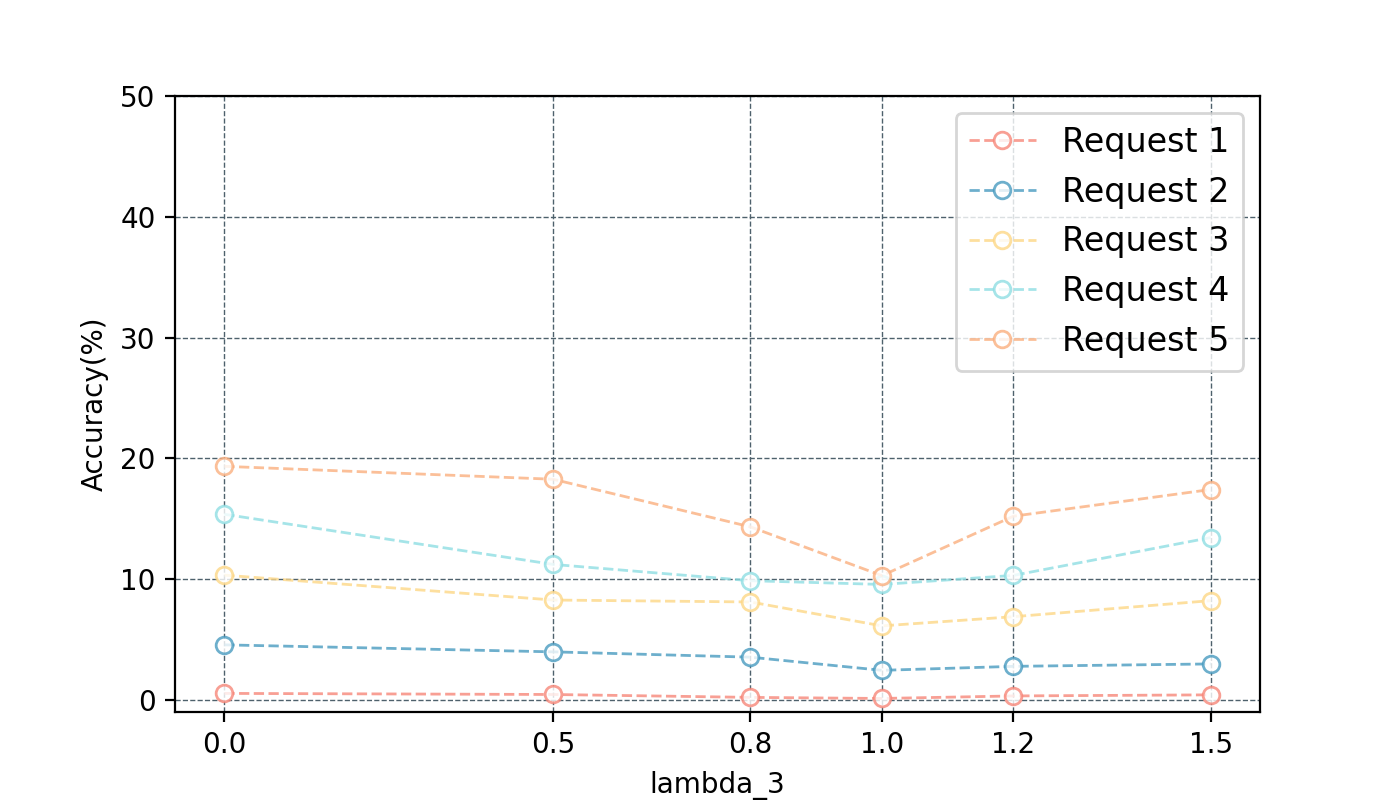}%
   \label{imagenetr-2}}
   \hfil
\subfloat[]{\includegraphics[width=1.8in]{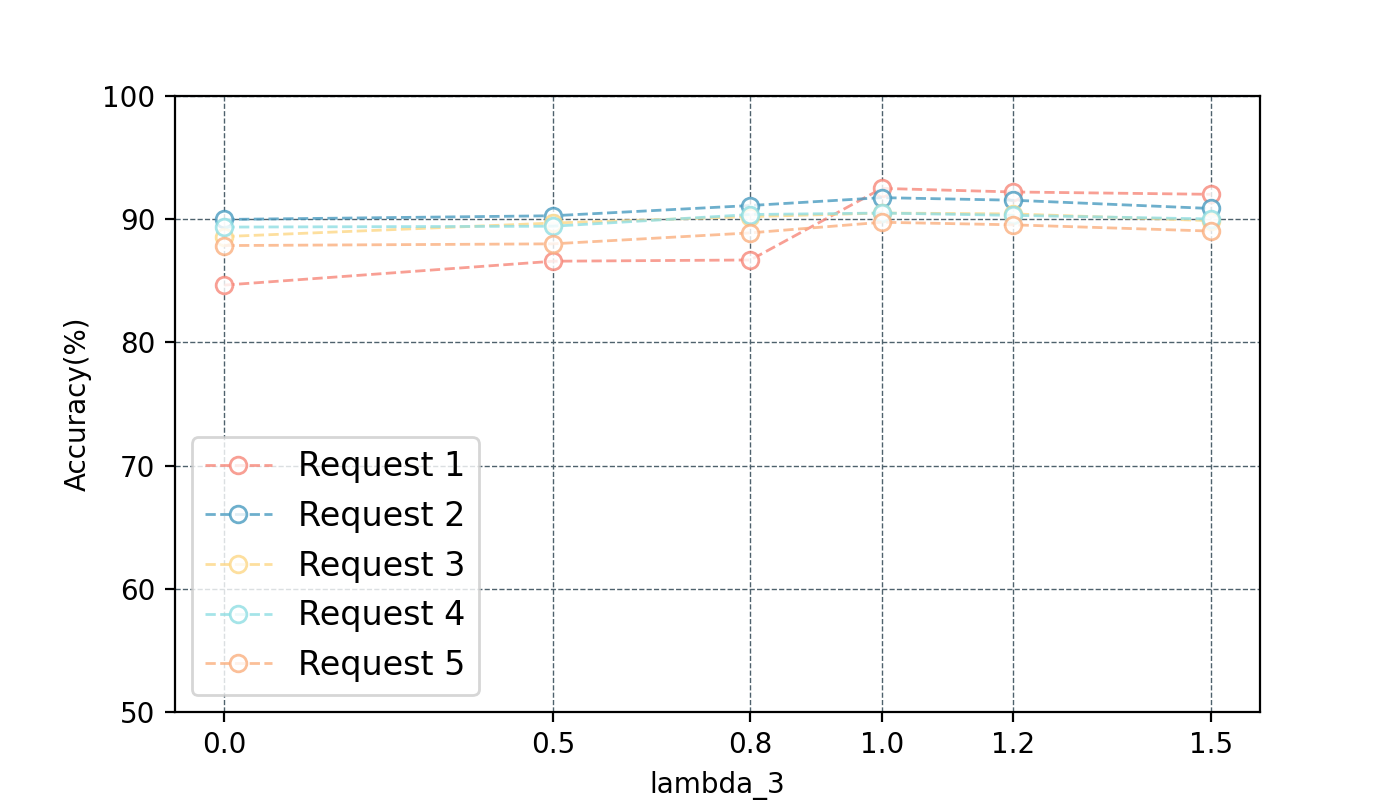}%
   \label{imagenetr-2}}
    \hfil
 \caption{Experimental results on the ScienceQA dataset.
 The left column shows the results of S.U., the middle column shows the results of D.U., and the right column shows the results of R.D..
 (a) (b) (c)The result of the hyperparameter $\lambda_1$.
    (d) (e) (f)The result of the hyperparameter $\lambda_2$(S.U.).
    (g) (h) (i)The result of the hyperparameter $\lambda_3$(S.U.).
    }
\label{lambda_12}
\end{figure*}

\begin{figure*}[h]
\vspace*{-12cm}
\centering
\subfloat[]{\includegraphics[width=1.8in]{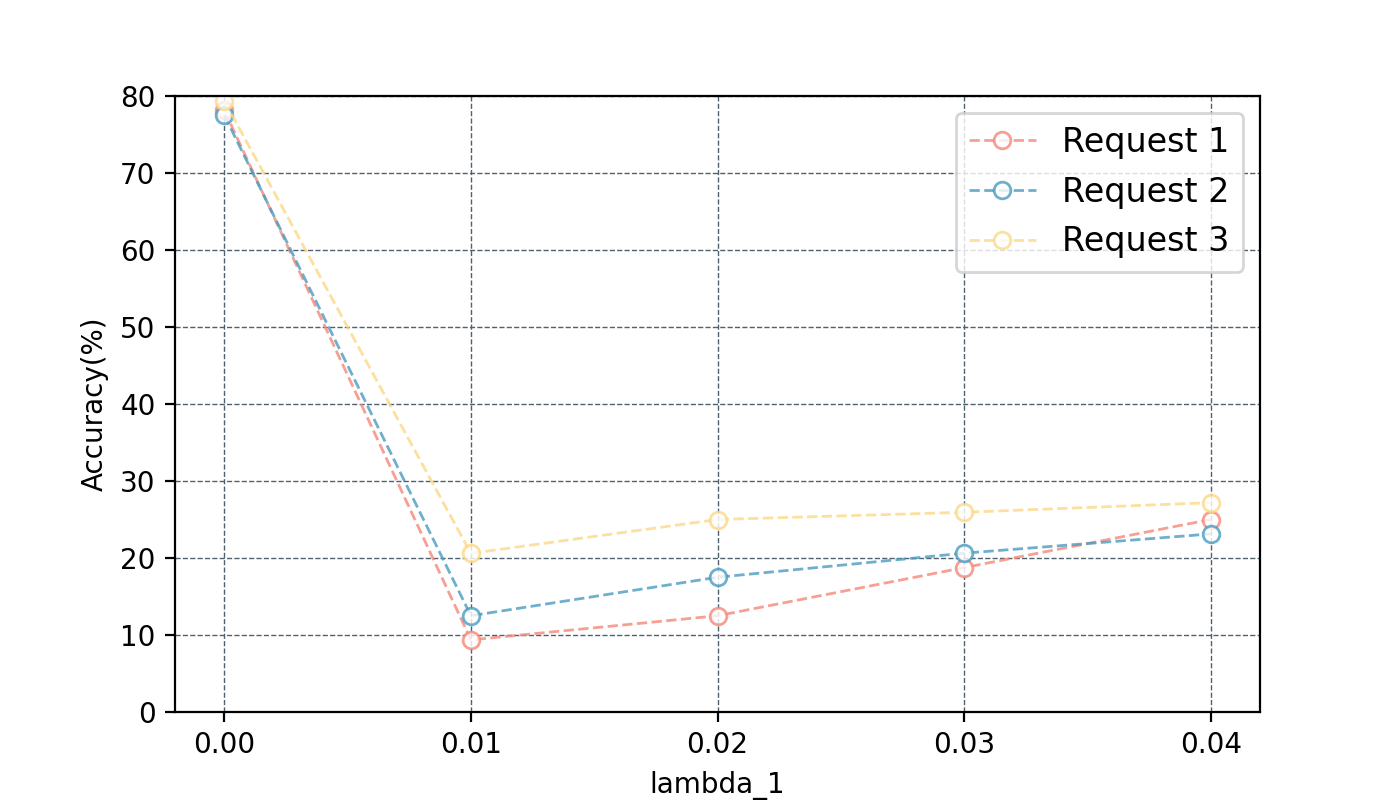}%
   \label{imagenetr-1}}
\hfil
\subfloat[]{\includegraphics[width=1.8in]{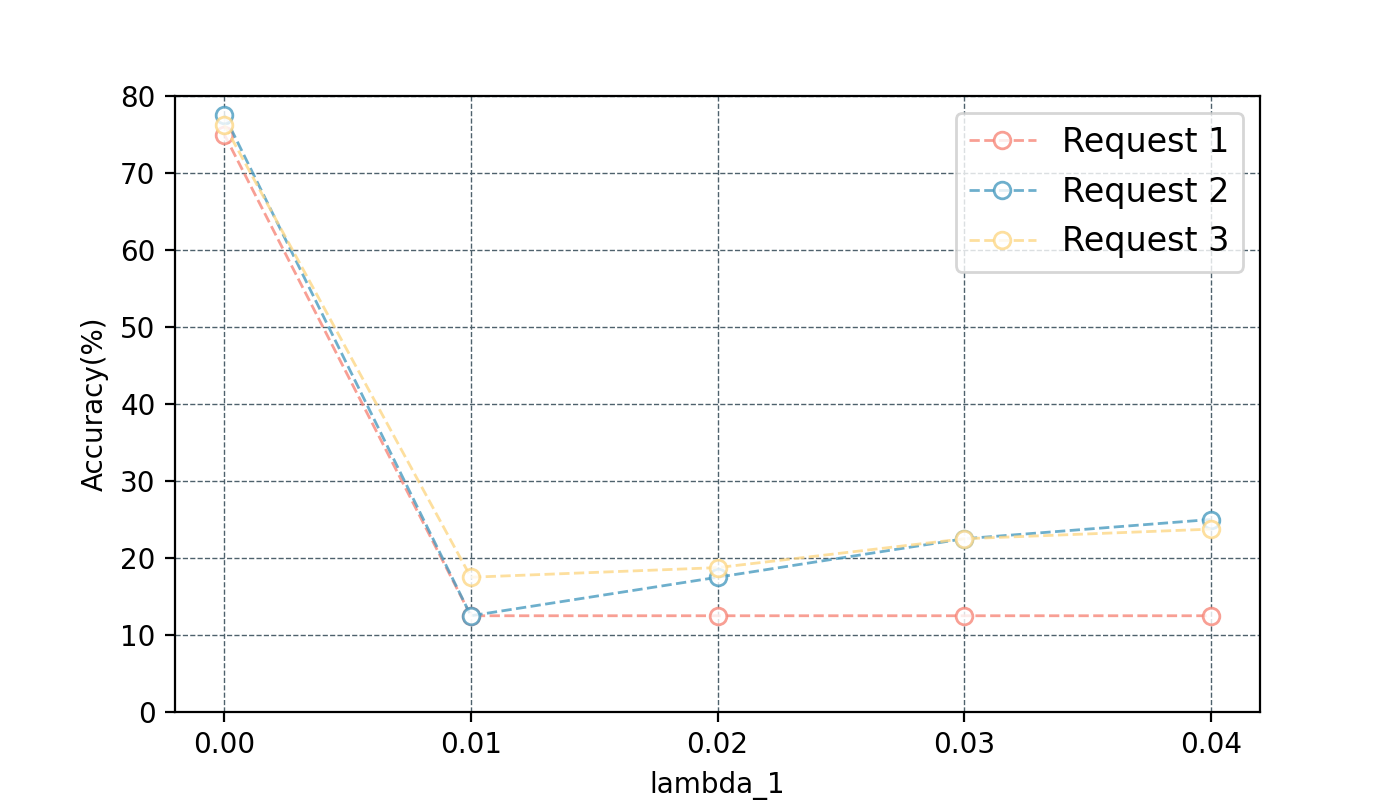}%
   \label{imagenetr-2}}
   \hfil
\subfloat[]{\includegraphics[width=1.8in]{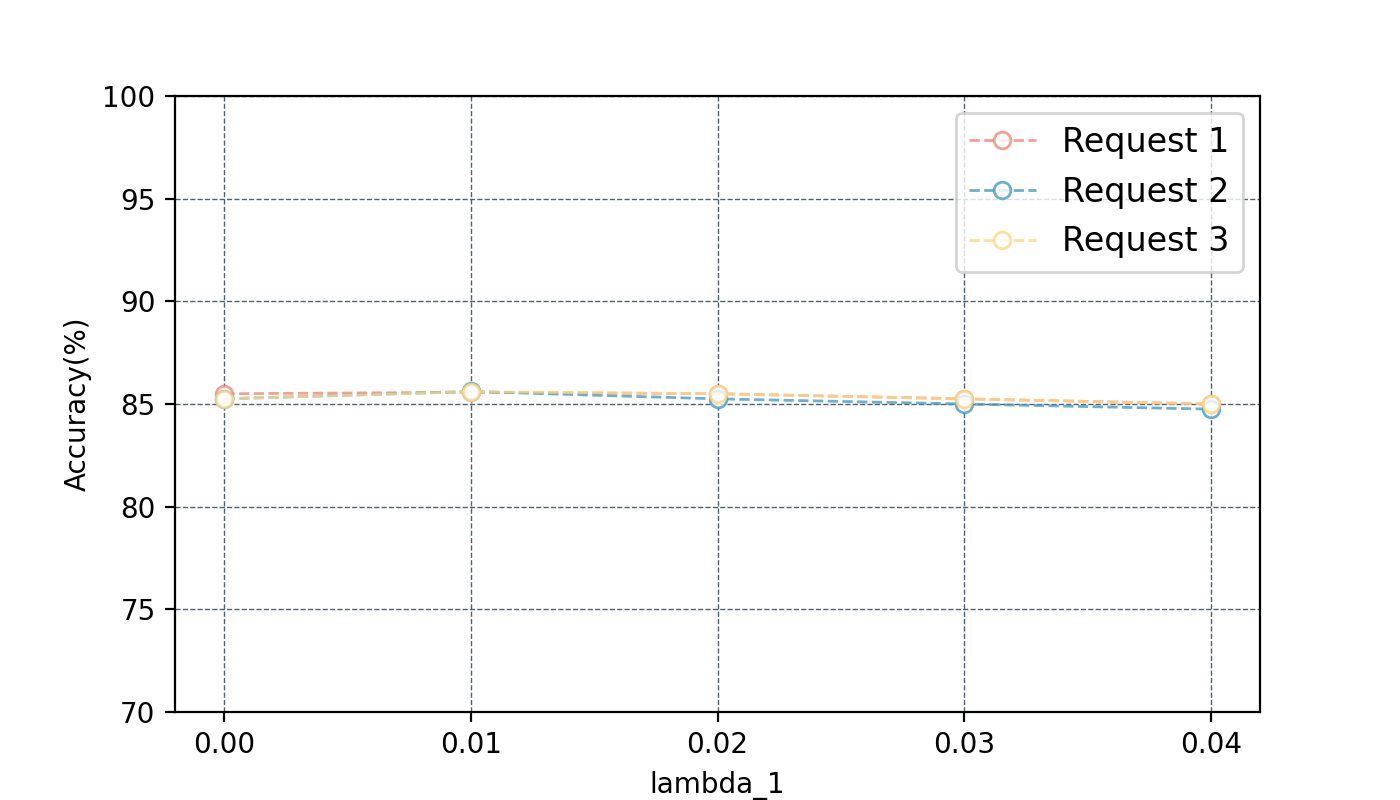}%
   \label{imagenetr-2}}
  \hfil
\subfloat[]{\includegraphics[width=1.8in]{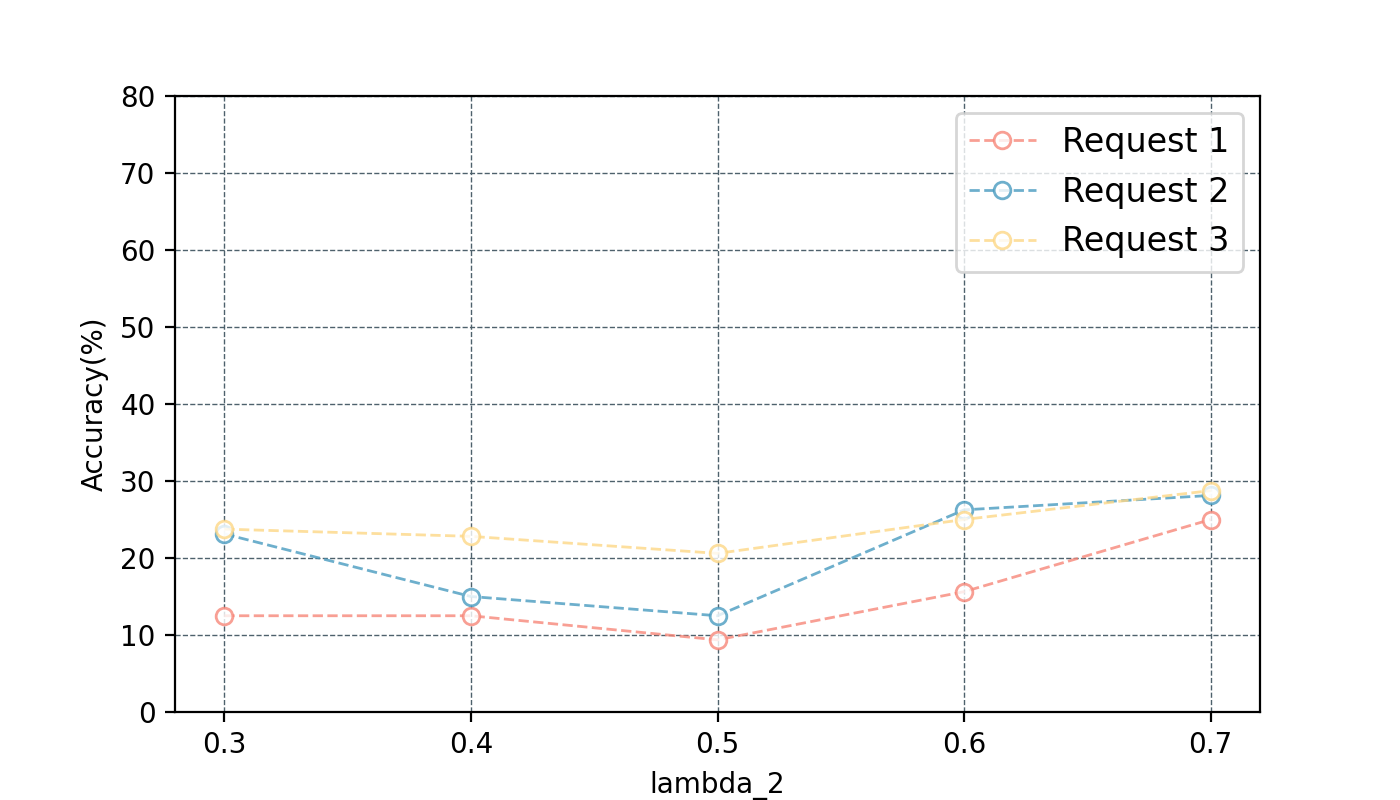}%
\label{imagenetr-2}}
  \hfil
\subfloat[]{\includegraphics[width=1.8in]{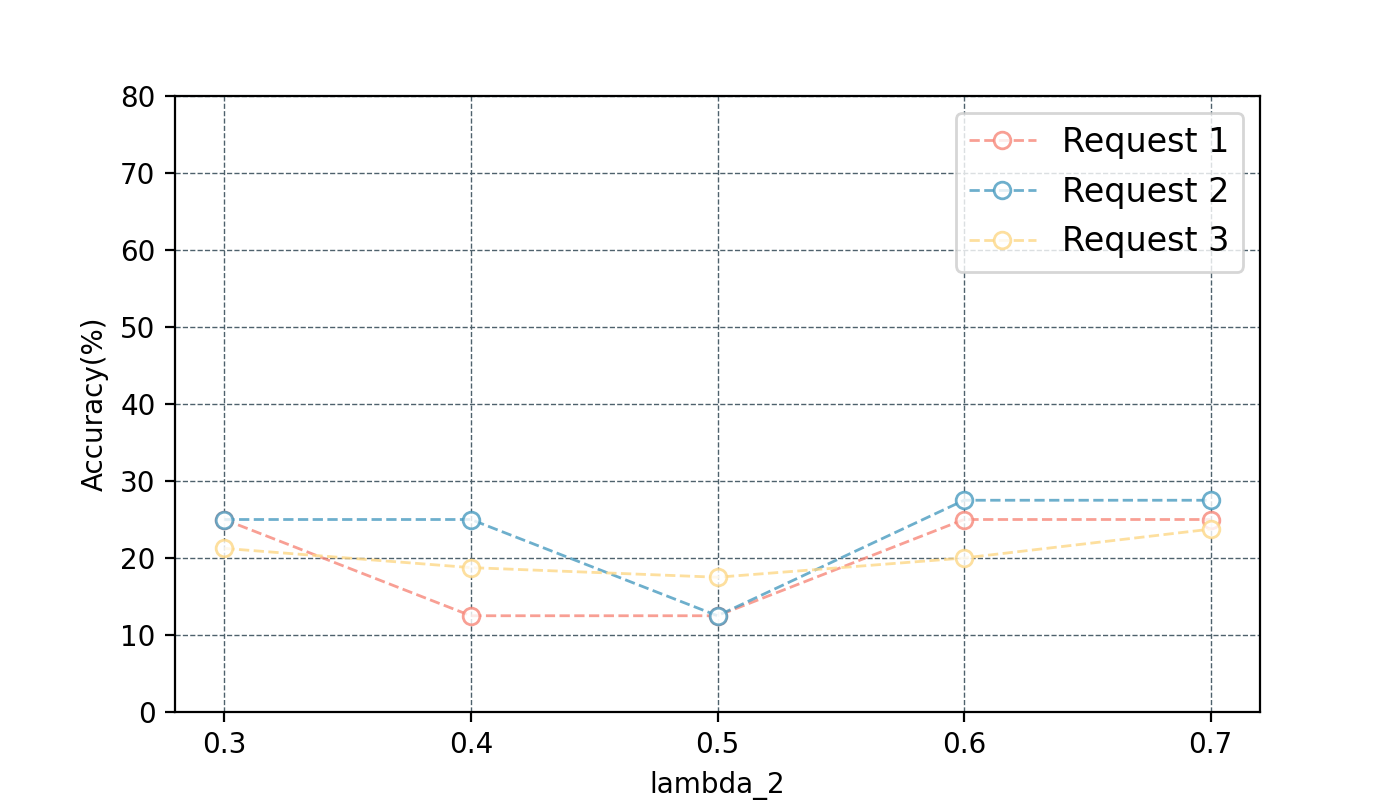}%
   \label{imagenetr-2}}
   \hfil
\subfloat[]{\includegraphics[width=1.8in]{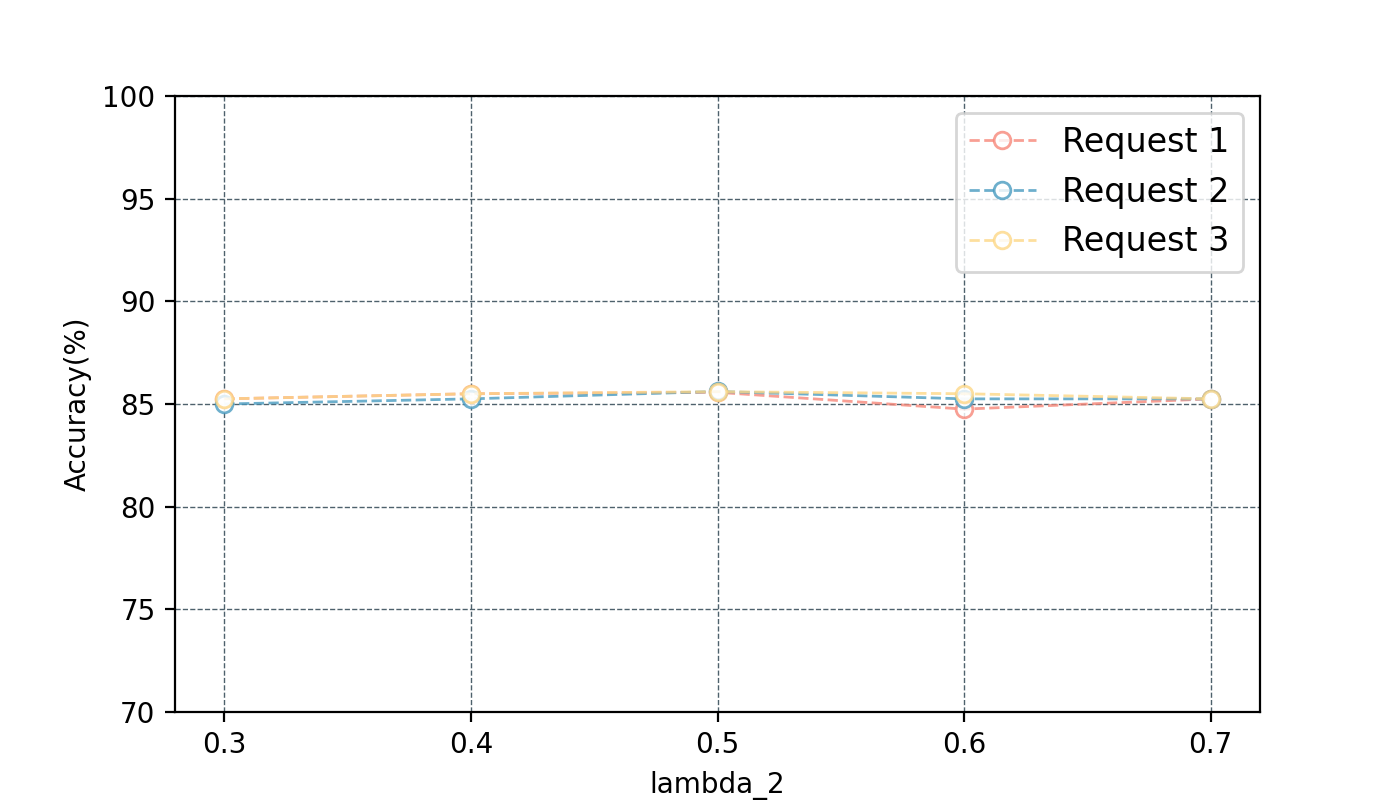}%
   \label{imagenetr-2}}
    \hfil
\subfloat[]{\includegraphics[width=1.8in]{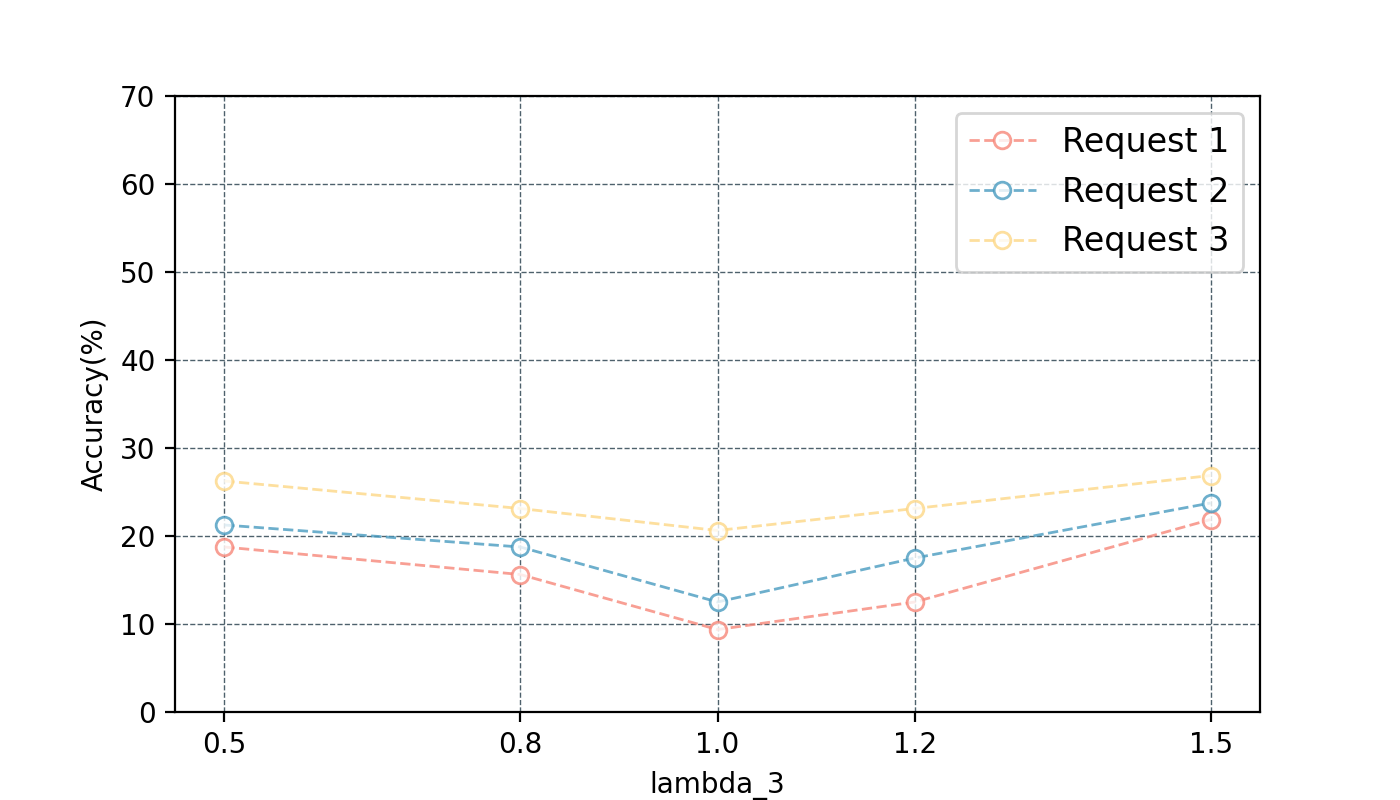}%
\label{imagenetr-2}}
  \hfil
\subfloat[]{\includegraphics[width=1.8in]{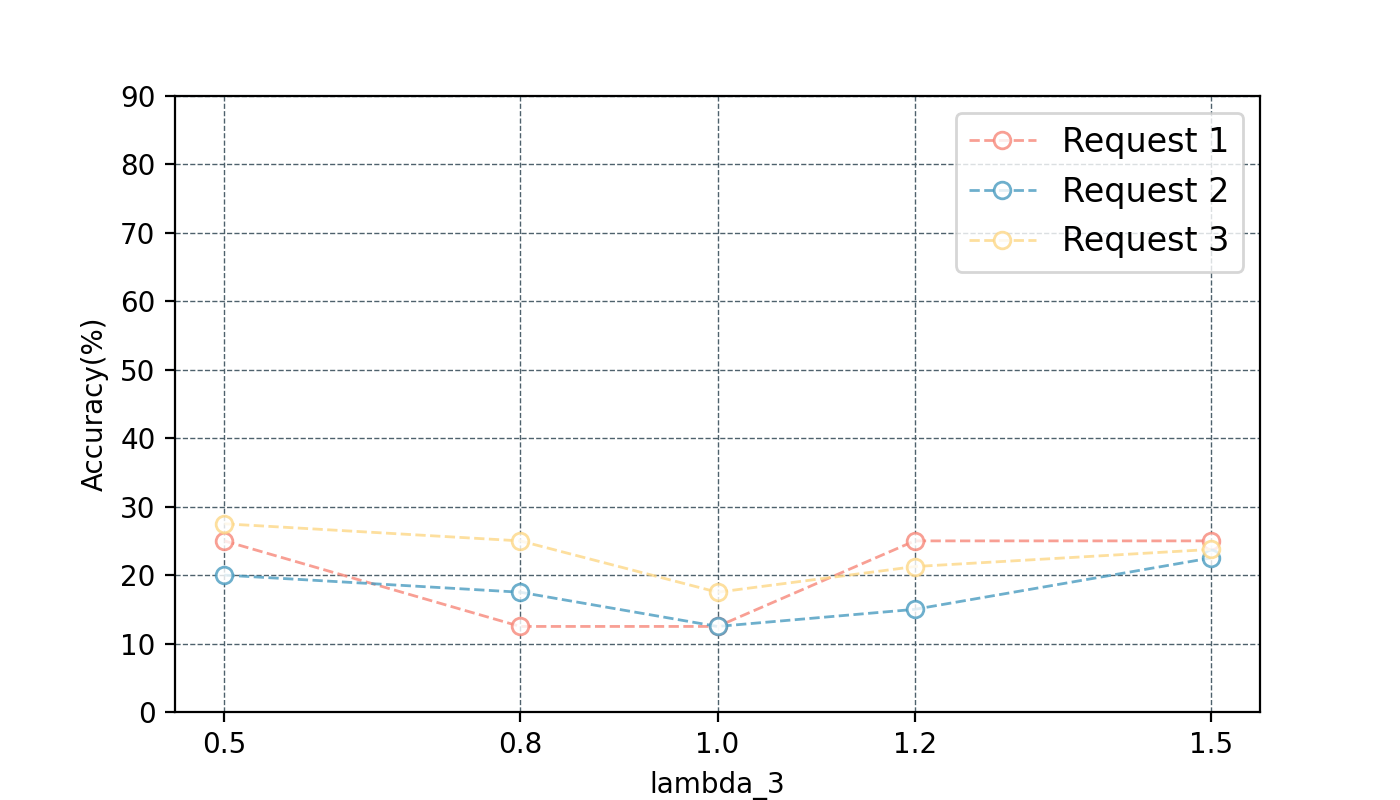}%
   \label{imagenetr-2}}
   \hfil
\subfloat[]{\includegraphics[width=1.8in]{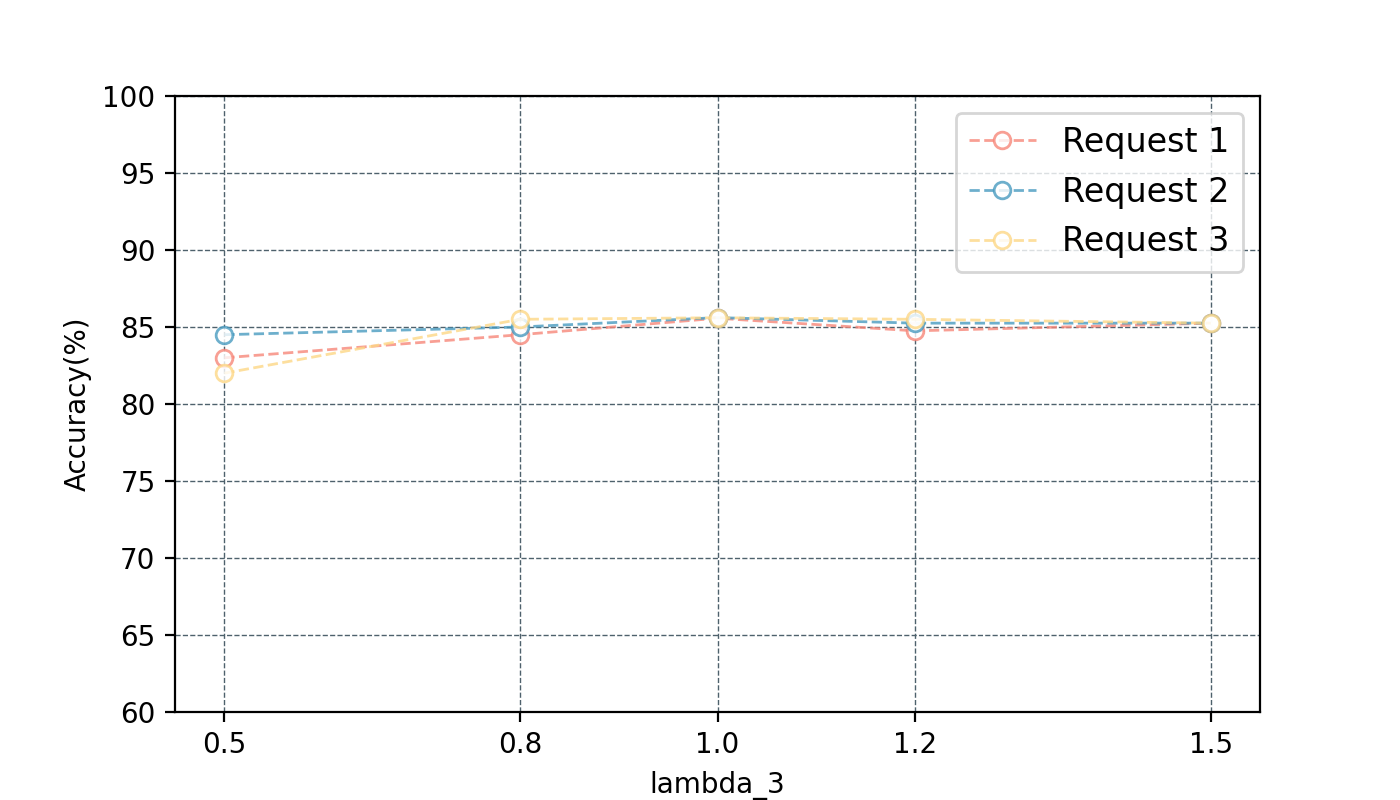}%
   \label{imagenetr-2}}
    \hfil
 \caption{Experimental results on the TOFU dataset.
  The left column shows the results of S.U., the middle column shows the results of D.U., and the right column shows the results of R.D..
 (a) (b) (c)The result of the hyperparameter $\lambda_1$.
    (d) (e) (f)The result of the hyperparameter $\lambda_2$(S.U.).
    (g) (h) (i)The result of the hyperparameter $\lambda_3$(S.U.).
    }
\label{lambda_tofu}
\end{figure*}

We conducted a comprehensive hyperparameter analysis on $\lambda_1$, $\lambda_2$, and $\lambda_3$ across two datasets. 
The results, presented in \autoref{lambda_12} and \autoref{lambda_tofu}, demonstrate that our chosen hyperparameter configuration consistently achieves optimal performance.

\subsection{Computation overhead analysis.}
During the reasoning process, the computational cost for LLM to update using the lora update paradigm proposed by us is 45.10 GFLOPs, the computational cost for OOD Detection is 709 MFLOPs, and the computational cost for the Distributional Shift Compensator is 13255 MFLOPs. 
The total computational cost is 59.064 GFLOPs.
In addition, our method requires less storage space. 
The storage space required by lora in $o^3$ is 39MB, while our method only requires 16MB, reducing the additional storage space requirement by 58\%.

\subsection{The comparison results of the Evaluation Indicators on the TOFU dataset.}
We choose the Truth Ratio and ROUGE-L as the indicators for evaluating the utility on the TOFU dataset.

The results of the Truth Ratio (\cite{tofu}) are presented in \autoref{TOFU-TR}. From these results, it can be seen that our method is superior to the comparison methods.
The formula for the Truth Ratio is as shown in \autoref{gongshi}.
\begin{align}
 R_{\text {truth }}=\frac{\frac{1}{\mathcal{A}_{\text {pert }}} \sum_{\hat{a} \in \mathcal{A}_{\text {pert }}} P(\hat{a} \mid q)^{1 /|\hat{a}|}}{P(\tilde{a} \mid q)^{1 /|\tilde{a}|}},
  \label{gongshi}
\end{align}
where $\tilde{a} $ is a paraphrased version of the correct original answer $a$, $\tilde{a} \in \mathcal{A} _{pert}$ are deliberately perturbed (incorrect) answers derived from $\tilde{a} $, and $\left | \tilde{a}  \right | $ denotes the number of tokens in $\tilde{a} $.

\begin{table*}[h]
 \caption{The results here are all the Truth Ratio (\cite{tofu}) corresponding to the aforementioned indicators.
 }

\renewcommand\arraystretch{1}
  \centering
  \resizebox{1\columnwidth}{!}{
  \begin{tabular}{c|ccccc|ccccc|ccccc|}
    \toprule
      Method & \multicolumn{5}{c}{Unlearning Request 1}& \multicolumn{5}{c}{Unlearning Request 2}& \multicolumn{5}{c}{Unlearning Request 3}\\
         Truth Ratio &  S.U.$\uparrow$ & D.U.$\uparrow$ & R.D.$\downarrow$ & R.A.$\downarrow$ & W.F.$\downarrow$ &  S.U.$\uparrow$ & D.U.$\uparrow$ & R.D.$\downarrow$ & R.A.$\downarrow$ & W.F.$\downarrow$ &  S.U.$\uparrow$ & D.U.$\uparrow$ & R.D.$\downarrow$ & R.A.$\downarrow$ & W.F.$\downarrow$ \\
    \midrule
    $o^3$  &0.74&0.66&0.57&0.54&1.54&0.65&0.65&0.57&0.54&1.54&0.66&0.65&0.57&0.54&1.54 \\
    \midrule
    \rowcolor{gray!20}
    Ours  &\textbf{1.00}&\textbf{1.11}&\textbf{0.56}&\textbf{0.54}&\textbf{1.22}&\textbf{0.99}&\textbf{0.97}&\textbf{0.56}&\textbf{0.54}&\textbf{1.22}& \textbf{0.97}&\textbf{1.01}&\textbf{0.56}&\textbf{0.54}&\textbf{1.22} \\
        
    \bottomrule
  \end{tabular}
  }
  \label{TOFU-TR}
\end{table*}



        

We compute the ROUGE-L recall score (\cite{tofu}), which acts as a surrogate for accuracy on the question answering task, as it accounts for the output phrasing to be slightly different than the ground truth.
The results of the ROUGE-L (\cite{tofu}) are presented in \autoref{TOFU-RL}. 
From these results, it can be seen that our method is superior to the comparison methods.

\begin{table*}[h]
 \caption{The results here are all the ROUGE-L (\cite{tofu}) corresponding to the aforementioned indicators.
 }

\renewcommand\arraystretch{1}
  \centering
  \resizebox{1\columnwidth}{!}{
  \begin{tabular}{c|ccccc|ccccc|ccccc|}
    \toprule
      Method & \multicolumn{5}{c}{Unlearning Request 1}& \multicolumn{5}{c}{Unlearning Request 2}& \multicolumn{5}{c}{Unlearning Request 3}\\
         ROUGE-L   &  S.U.$\downarrow$ & D.U.$\downarrow$ & R.D.$\uparrow$ & R.A.$\uparrow$ & W.F.$\uparrow$ &  S.U.$\downarrow$ & D.U.$\downarrow$ & R.D.$\uparrow$ & R.A.$\uparrow$ & W.F.$\uparrow$ &  S.U.$\downarrow$ & D.U.$\downarrow$ & R.D.$\uparrow$ & R.A.$\uparrow$ & W.F.$\uparrow$ \\
    \midrule
    $o^3$  &0.0771&0.0627&0.9675&0.9330&0.8960&0.1939&0.1255&0.9677&0.9330&0.8960&0.1843&0.5189&0.9675&0.9330&0.8960 \\
    \midrule
    \rowcolor{gray!20}
    Ours  &\textbf{0.0425}&\textbf{0.0365}&\textbf{0.9675}&\textbf{0.9330}&\textbf{0.9083}&\textbf{0.1333}&\textbf{0.1044}&\textbf{0.9683}&\textbf{0.9330}&\textbf{0.9083}&\textbf{0.1322}&\textbf{0.1062}&\textbf{0.9683}&\textbf{0.9330}&\textbf{0.9083} \\
        
    \bottomrule
  \end{tabular}
  }
  \label{TOFU-RL}
\end{table*}

\end{document}